\documentclass[journal]{IEEEtran}

\usepackage{xcolor,soul,framed} %,caption

\colorlet{shadecolor}{yellow}
\usepackage[pdftex]{graphicx}
\graphicspath{{../pdf/}{../jpeg/}}
\DeclareGraphicsExtensions{.pdf,.jpeg,.png}

\usepackage[cmex10]{amsmath}
\usepackage{array}
\usepackage{mdwmath}
\usepackage{mdwtab}
\usepackage{eqparbox}
\usepackage{url}
\usepackage{amsthm,amssymb}
\usepackage{graphicx}
\usepackage{multirow}

\begin{document}

  \title{Relation Learning and Aggregate-attention for Multi-person Motion Prediction}
  \author{Kehua Qu†, Rui Ding†*, Jin Tang* % stops a space

  \thanks{Kehua Qu, and Rui Ding are with Information Engineering College, Capital Normal University, Beijing 100048, China. Email: 2221002103@cnu.cn, 5758@cnu.cn.}
  \thanks{Jin Tang is with the School of Intelligent Engineering and Automation, Beijing University of Posts and Telecommunications, Beijing 100876, China. E-mail: tangjin@bupt.edu.cn.}% <-this % stops a space
  \thanks{*Corresponding authors.}%
  \thanks{†These authors contributed equally to this work.}
}

\markboth{IEEE TRANSACTIONS ON MULTIMEDIA
}{Roberg \MakeLowercase{\textit{et al.}}: High-Efficiency Diode and Transistor Rectifiers}

% ====================================================================
\maketitle

% === ABSTRACT ====================================================================
% =================================================================================
\begin{abstract}
Multi-person motion prediction is an emerging and intricate task with broad real-world applications. Unlike single person
motion prediction, it considers not just the skeleton structures or human trajectories but also the interactions between others. Previous methods use various networks to achieve
impressive predictions but often overlook that the joints relations within an individual (intra-relation) and interactions among groups (inter-relation) are distinct types of representations. These methods often lack explicit representation of inter\&intra-relations,
and inevitably introduce undesired dependencies. To address this issue, we introduce a new collaborative framework for multi-person motion prediction that explicitly modeling these relations: a GCN-based network for intra-relations and a novel reasoning network for inter-relations. Moreover, we propose a novel plug-and-play aggregation module called the Interaction Aggregation Module (IAM), which employs an aggregate-attention mechanism to seamlessly integrate these relations. Experiments indicate that the module can also be applied to other dual-path models. Extensive experiments on the 3DPW, 3DPW-RC, CMU-Mocap, MuPoTS-3D, as well as synthesized datasets Mix1 \& Mix2 (9$\sim$15 persons), demonstrate that our method achieves state-of-the-art performance.
\end{abstract}

% === KEYWORDS ====================================================================
% =================================================================================

\section{Introduction}
Human motion prediction (HMP) aims to predict future human motion sequences based on the observed sequences. The HMP plays a major role in many real-world applications, such as autonomous driving \cite{tang2023collaborative,fang2023tbp}, robotics \cite{gao2021human,8460651}, surveillance systems \cite{xu2022remember}. Due to the remarkable development of deep learning, the HMP has made unprecedented progress in recent years \cite{zhong2022spatio,nie2023triplet,10064318,wang2023dynamic}. However, single-person motion prediction methods focus solely on intra-relations, which contain the relative positions and movement patterns among body joints (e.g., hip, ankle, wrist) of the same individual. Humans are intuitively social agents, and they continuously interact with other people and their motion may also be influenced by the motions of others. Therefore, multi-person motion prediction task carries more practical significance compared to single-person motion prediction. This task is more challenging because of the sophisticated interactions across different individuals.

\begin{figure}
	
	\centering
	
	\includegraphics[width=0.5\textwidth]{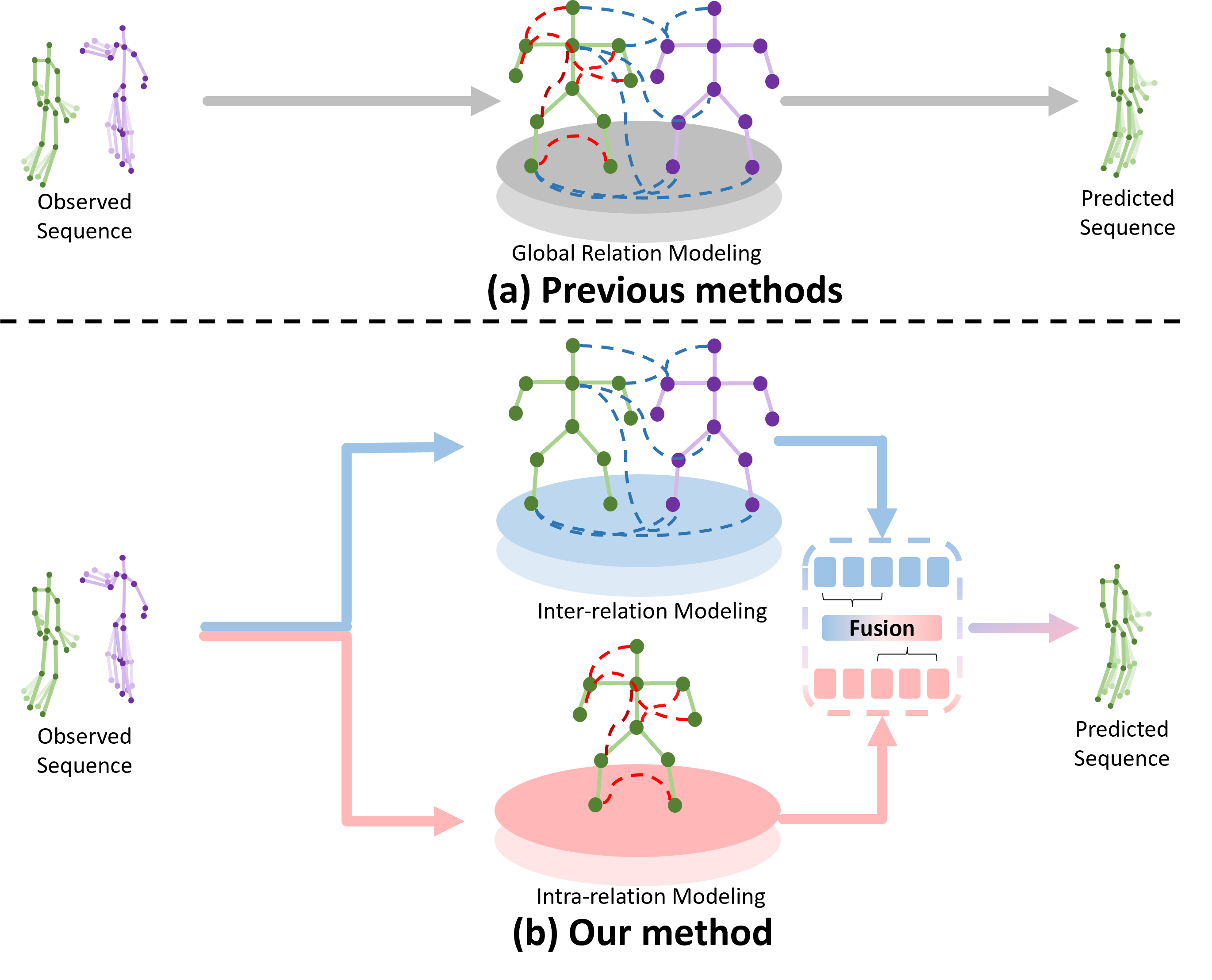}

	\caption{Compared to previous methods \cite{wang2021multi,9709907,xu2023joint,10194334} with relation learning, we propose a new collaborative learning framework
that explicitly explore joints relations, including intra-relations and inter-relations. The red and blue dashed lines indicate the inter-relation and intra-relation, respectively.}
	
	\label{intro}
  
\end{figure}

\begin{figure*}
	
	\centering
	
	\includegraphics[width=1\textwidth]{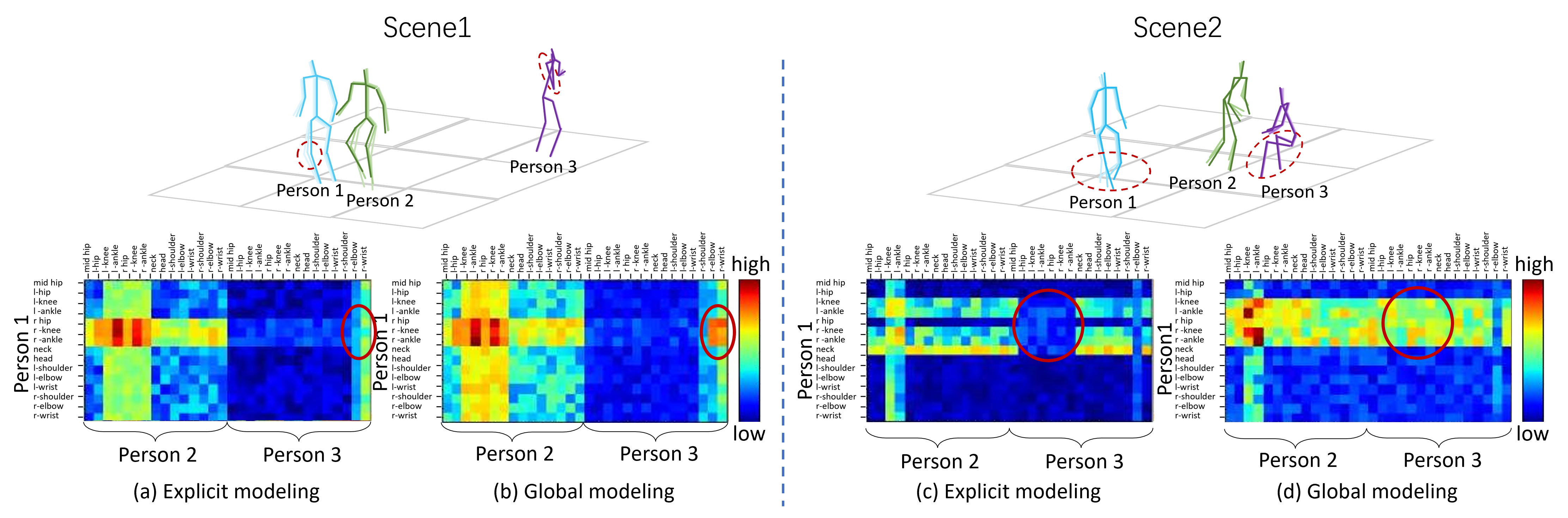}

	\caption{Visualization of Pearson correlation coefficient (PCC) between different individuals. We conducted two Transformer-based architecture experiments on the CMU-Mocap dataset: (i) The explicit relation modeling adopts cross-attention to learn inter-relations between different individuals’ joints, and self-attention to learn intra-relation of each individual’s joints. (ii) The global modeling utilizes self-attention to learn all relations of all inputting skeleton joints. For each scene, the upper image shows the true scene in the sequence. The lower image shows the visualization of PCC between the 15 joints of person 1 and the other two persons’ joints. The red color indicates higher correlation (larger PCC) between two joints, while the blue indicates lower (smaller PCC).}
	\label{intro2}

\end{figure*}

The previous works on multi-person motion prediction generally tend to exploring interactive behaviors by global relation modeling in which all skeleton joints are treated as a whole to establish the relations between them via self-attention \cite{wang2021multi,9709907,xu2023joint,10194334}. As shown in Fig. \ref{intro}(a), these methods ignore the distinct representations of inter-relation and intra-relation. Noted that we define inter-relation as the relation of across different individuals’ joints, as the blue dash line shown. Corresponding, we define intra-relation as the relations of an individual’s joints, as the red dash line shown. By inputting all skeleton joints as a whole into the network, these methods may inevitably introduce undesired relations which weakens the learning effects of interaction information and simplifies the constraints of joints. 

To reveal the advantages of explicit representation relations, we conducted some Transformer-based architecture experiments on the CMU-Mocap dataset: (i) The explicit relation modeling adopts cross-attention to learn inter-relations between different individuals’ joints, and self-attention to learn intra-relation of each individual’s joints. (ii) The global modeling utilizes self-attention to learn all relations of all inputting skeleton joints. Both baselines are based on Transformer architecture. Inspired by methods \cite{cao2022pkd,zhu2023ipcc}, we employ the Pearson correlation coefficient (PCC) to model the correlation between different people in the observed scene. The calculation can be expressed as follow:

\begin{equation}
\begin{aligned}
\mathbf{PPC\left(P_{1},P_{2}\right)} = & \frac{\sum_{t=1}^{T} \left( \mathbf{P_{1}}\left(t,j\right) - \overline{P}_{1}\left(j\right) \right)} {\sqrt{\sum_{t=1}^{T} \left( \mathbf{P_{1}}\left(t,j\right) - \overline{P}_{1}\left(j\right) \right)^{2}}} \\
   & \times \frac{\sum_{t=1}^{T} \left( \mathbf{P_{2}}\left(t,j\right) - \overline{P}_{2}\left(j\right) \right)} {\sqrt{\sum_{t=1}^{T} \left( \mathbf{P_{2}}\left(t,j\right) - \overline{P}_{2}\left(j\right) \right)^{2}}}
\end{aligned}
\end{equation}

Where $\mathbf{P_{n}}$ denote the motion representation of $n$-th person. $t$ denotes the $t$-th motion sequence and $j$ denotes the $j$-th skeleton joint. $\overline{P}_{n}$ denotes the mean value of person n. We visualize PCC between different individuals' joints from two scenes, as shown in Fig. \ref{intro2}. In the first scene, person 1 is walking with person 2. Meanwhile, person 3 is phoning by moving his right arm (in red dashed circle). The red solid circled portions of Fig. \ref{intro2}(a) and Fig. \ref{intro2}(b) indicate the correlation between person 3's right hand and person 1's right leg. (The closer the color is to red, the greater the correlation.) Surprisingly, using the global relation modeling approach, person 3’s right hand is correlated with person 1’s leg joints, which is not consistent with the fact, as shown in Fig. \ref{intro2}(b). In the second scene, person 2 is conversing with a seated person 3 while person 1 is passing by them. In the observed sequence, there is little correlation between the trajectory of person 1 and the lower body of the stationary person 3 (in red dashed circle). From Fig. \ref{intro2}(c), we can see that the response values between the lower body (hip, knee, ankle) of person 1 and the lower body of person 3 are lower (within the red circle) than those of other body parts. The phenomenons demonstrate that the explicit modeling is more effective represent relations and avoid the undesired one. Moreover, global modeling methods make it difficult to assess which parts of the relationships are crucial for prediction accuracy, compromising the model's interpretability.

Therefore, we advocate explicitly modeling relations, as shown in Fig. \ref{intro}(b). Specifically, we propose a new explicitly modeling framework that contains Graph Convolutional Networks (GCN) for intra-relations and the cross-attention attention mechanism for inter-relations. GCN-based methods have surpassed attention mechanism in single-person motion prediction tasks, as demonstrated by references \cite{zhong2022spatio, sofianos2021space, dang2021msr}, effectively isolating these influences from own joints. In contrast,  attention-based approaches excel at handling complex dependencies \cite{lin2022survey} and prove more effective in multi-person scenarios \cite{wang2021multi, vendrow2022somoformer, xu2023joint}. In particular, the cross-attention mechanism allows us to exclusively focus on the influence of others (inter-relations) while extracting interaction features, effectively isolating these influences from own joints.

\begin{figure}
	
	\centering
	
	\includegraphics[width=0.45\textwidth]{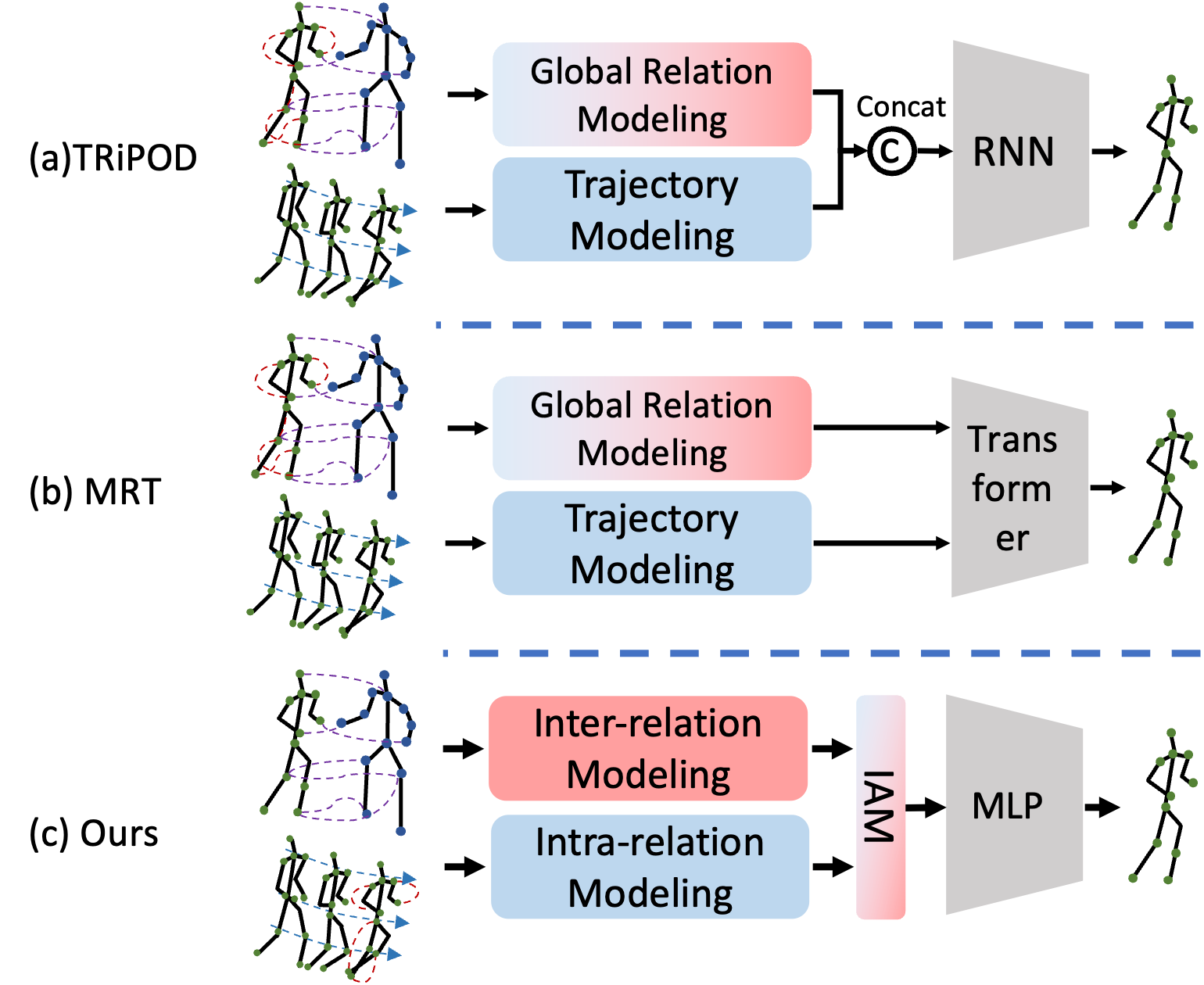}

	\caption{The illustration of different fusion strategies: (a) TRiPOD \cite{9709907} concats features from two different branch, then feeds them to an RNN decoder. (b) MRT \cite{wang2021multi} feeds distinct features to a Transformer decoder to explore their dependency automatically. (c) Our method leverages inter\&intra-relations by a designed fusion module (IAM).}
	
	\label{intro3}
  
\end{figure}

Furthermore, the fusion of intra-relation and inter-relation is critical for high-quality prediction results. Since the simple addition or concatenation operation only mixes the features according to the dimensions \cite{xu2023joint,9709907}, it includes too much irrelevant information to fully extract more helpful information. To tackle the fusion issues, comparing to the previous approaches as shown in Fig. \ref{intro3}, we introduce a novel plug-and-play aggregation module called the Interaction Aggregation Module (IAM), which utilizes a variant attention mechanism that allows one type of relation to directly influence another, ensuring that the fused feature fully takes both types of relationships into account. Moreover, we conduct experiments on several baselines with replacing the our IAM and all these methods achieve improved results. This approach not only improves predictions but also showcases IAM’s robust plug-and-play capabilities, as confirmed by our experimental results.

We perform our experiments on multiple datasets, including 3DPW \cite{Marcard_2018_ECCV}, 3DPW-RC \cite{xu2023joint}, CMU-Mocap \cite{cmumocap2003}, MuPoTS-3D \cite{mehta2018single}, and synthesized datasets Mix1\&Mix2 \cite{wang2021multi} (9$\sim$15 persons). The quantitative results demonstrate that our method achieves state-of-the-art performance.

In summary, our contributions are as follows:
\begin{itemize}
\item We propose a collaborative learning framework for multi-person motion prediction to explicitly modeling both the intra-relations within individuals and the inter-relations between them. 
Explicit modeling clearly illustrate the role of these two relations, enhancing the model's robustness and interpretability.

\item We propose a novel plug-and-play aggregation module called the Interaction Aggregation Module (IAM), which introduces an aggregate-attention mechanism to seamlessly integrate intra\&inter-relations. Experiments show that this module effectively aggregates these relations and is adaptable to other models requiring integrated information.
\item We conduct experiments on several common datasets and our method achieves state-of-the-art performance on all datasets. 
\end{itemize}

\section{Related Work}
\subsection{Single-Person Motion Prediction.}
Early research in single-person motion prediction focused on time series models \cite{ghosh2017learning,gopalakrishnan2019neural,wang2021pvred}. Wang et al. \cite{wang2021pvred} propose a novel Position-Velocity Recurrent Encoder-Decoder that effectively utilizes pose velocities and temporal positional information. Despite their success, it seems inappropriate to consider human motion purely as a time series. Human motion involves multiple joint points, each with inherent spatial relationships. For example, the motion relationship between the knee and the ankle differs from that between the elbow and the shoulder. These spatial relationships cannot be captured separately in a time series. To address this issue, GCN-based models are now widely used \cite{zhong2022spatio,sofianos2021space,dang2021msr,wang2023dynamic}. Sofianos et al. \cite{sofianos2021space} propose a classic model based on GCN that integrates joint spatio-temporal features. This approach enhances the interaction between body joint connections and their temporal motion patterns, facilitating more effective modeling of human dynamics. Recently, Transformer-based methods have gained widespread attention in the field of motion prediction \cite{mao2020history, aksan2021spatio, nie2023triplet,10064318,10306327}. Specifically, Aksan et al. \cite{aksan2021spatio} employ a dual-stream architecture based on the Transformer to address both spatial and temporal modeling. However, these methods primarily address single-person scenarios and are unsatisfactory for multi-person tasks.

\begin{figure*}[t]
	
	\centering
	
	\includegraphics[width=1\textwidth]{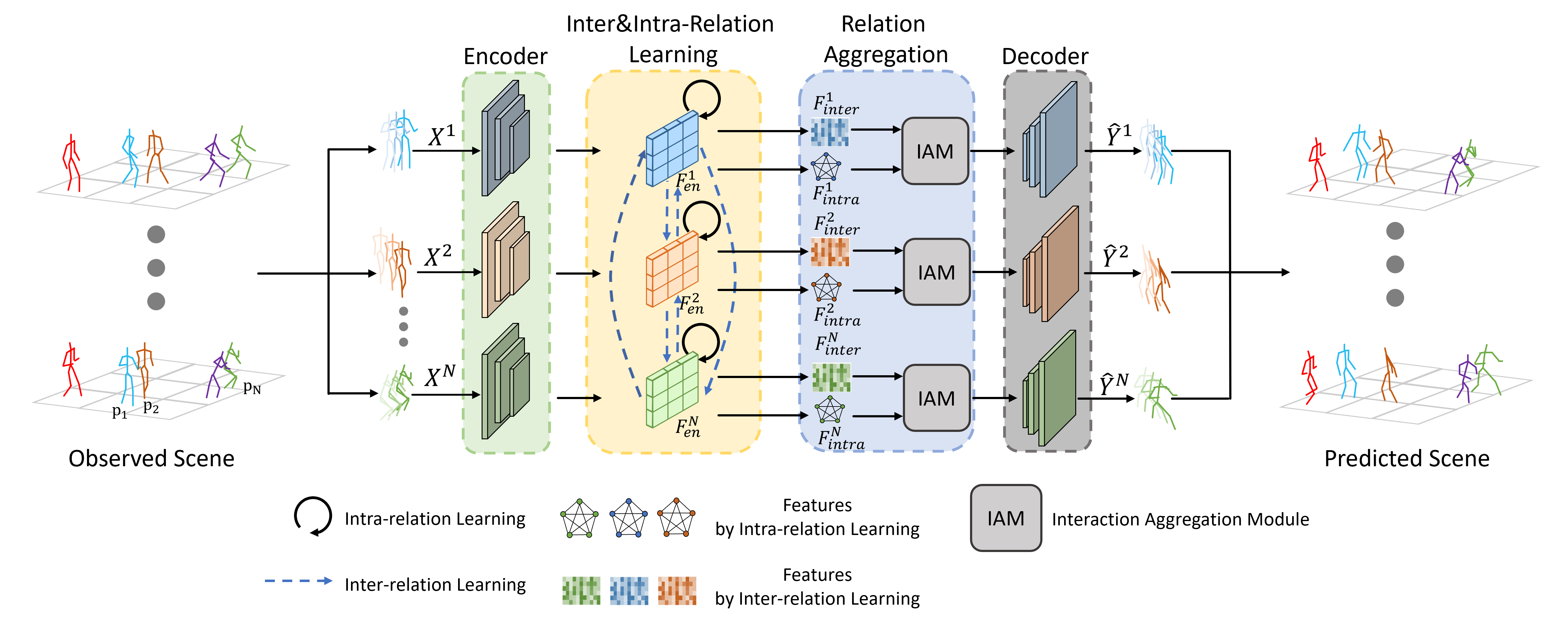}

	\caption{The architecture of our framework. The method contains : i)Encoder, ii) Intra\&inter-relation learning,
iii) Relation aggregation, iv)Decoder.}
	\label{arc}

\end{figure*}

\subsection{Human-Human/Object Interaction.} 
Humans are intuitively social agents, they continuously interact with other people and objects. Modeling
such interactions has been
proven to be effective in various tasks such as 3D human pose estimation \cite{li2022exploiting,9157826, zou2023snipper}, human action recognition \cite{10001762,liu2023temporal,9145701}, and human/object trajectory prediction \cite{9010834,10.1007/978-3-030-58452-8_23,9795092}.

Cao et al. \cite{10.1007/978-3-030-58452-8_23} incorporate scene context (human-object) into motion prediction but overlook human-human interactions, which are more complex due to the dynamic nature of humans compared to static objects.
Corona et al. \cite{9157826} expand the utilization of interactive information in motion prediction, but only limited human-human or human-object correlations are modeled due to the traditional time-series networks. Adeli et al. \cite{9145701}
develop a social context-aware motion prediction
framework to capture the interaction between human and human, 
but the social pooling of method eliminates the characterization of different people's actions. Nevertheless, while these models can capture human-human/object interactions, they cannot be directly applied to HMP due to objectives.

\subsection{Multi-Person Motion Prediction.} 
Obviously, it does not make sense to divide a multi-person task into multiple single-person tasks \cite{parsaeifard2021learning,saadat2021towards}. 
Adeli \cite{9709907} is the first to consider capturing spatial relations of all persons. Subsequently, more methods \cite{wang2021multi, vendrow2022somoformer, xu2023joint,10194334} devote to considering the human interaction.
Parsaeifard et al. \cite{parsaeifard2021learning} use a traditional LSTM \cite{6795963} encoder-decoder to predict human trajectories, but this approach does not account for crucial interactions between human actions. Adeli et al. \cite{9709907} adopt GAT \cite{veličković2018graph} to model both people and objects as graph nodes to capture their interactions. Since the attention mechanism of GAT only acts on direct neighbor nodes, the model can capture only local features at each aggregation, making it difficult to directly access global features. This limitation can result in unsatisfactory predictions when global context is crucial. Nowadays, more methods \cite{wang2021multi, vendrow2022somoformer, xu2023joint, 10194334, guo2022multi} demonstrate the effectiveness of Transformer in this task, as each node can directly focus on all other nodes, not just its neighbors. This capability allows the model to capture global features. Despite these advancements, there is potential for further improvement. Wang et al. \cite{wang2021multi} employ a local-range Transformer to encode the motion of an individual in the sequence and a global-range Transformer to encode the motion of multiple individuals. Both encoded motions are then sent to a Transformer-based decoder to predict future motion. Guo et al. \cite{guo2022multi} proposes a novel cross interaction
attention mechanism that exploits historical information of
both persons, and learns to predict cross dependencies between
the two pose sequences. This method primarily focuses on capturing human-to-human interactions and ignore the temporal features modeling, which are equally crucial for accurate prediction. Edward et al. \cite{vendrow2022somoformer} model the interaction between all joints by utilizing the Transformer’s attention mechanism. Liu et al. \cite{10194334} introduce two novel modules to model human trajectories and skeleton joints separately, yet result in the absence of spatio-temporal integration. Xu et al. \cite{xu2023joint} introduces physical constraints into predictions, indicating that joints connected by bone show stronger associations and that joints belonging to the same individual are related. The above methods treat individuals in a scene collectively when modeling interactions between persons, ignore the distinction between the spatial relationships within a person and those between multiple persons. To address this issue, we introduce a novel framework for capturing intra\&inter-relations explicitly.

\section{Methodology}

We design a novel and explicit framework for multi-person motion prediction, including encoder-decoder, inter/intra-relation learning and relation aggregation, as shown in Fig. \ref{arc}.

\subsection{Overview}
 Given a scene with $N$ persons, each has $J$ skeleton joints, we define the observed sequence of the $n$-th person as ${X^{n} }=\left \{ x_{1}^{n} , x_{2}^{n},...x_{T}^{n} \right \} $, where $n\in  \left \{ 1,2,...,N \right \}$, $N$ denotes the number of observed people, $T$ represents the observed motion sequence length, and each $x_{t}^{n} = (j_{t,1}^{n},j_{t,2}^{n},...,j_{t,J}^{n})\in \mathbb{R}^{J\times 3}$ denotes the
joints' 3-D coordinates of the $n$-th person
at the $t$-th motion sequence. Our objective is to predict the future motion sequence of the $n$-th person, denoted as ${\widehat{Y}^{n} }=\left \{ \widehat{y}_{T+1}^{n} , \widehat{y}_{T+2}^{n},...,\widehat{y}_{T+P}^{n} \right \} $, using the observed sequence $X^{n} $, where $P$ denotes the predicted motion sequence length. The ground-truth of the $n$-th person can be defined as ${Y^{n} }=\left \{ y_{T+1}^{n} , y_{T+2}^{n},...,y_{T+P}^{n} \right \}$. 
\subsection{Encoder}
Instead of using joint position coordinates as input, we utilize the position differences between consecutive time steps to represent individual velocities, thereby augmenting the joint information. Each velocity $v_{t}^{n}$ is thus expressed as follows:

\begin{align}
     v_{t}^{n}= \left\{\begin{matrix}
0\ \ \ &(t=1)\\
x_{t}^{n}-x_{t-1}^{n}\ \ \ &(2\leqslant t\leqslant T)\\  
\end{matrix}\right.
\end{align}
Following this approach, we obtain a new sequence ${V^{n} }=\left \{ v_{1}^{n} , v_{2}^{n},...v_{T}^{n} \right \} $ as the input of the encoder. The output $F_{en}^{n}$ of encoder is as follows:
\begin{equation}
    F_{en}^{n} = \mathbf{ MLP}(V^{n}) 
\end{equation}
The MLP block comprises three linear layers, an activation function, and a dropout layer, designed to expand the feature dimensions.

\subsection{Inter/Intra-Relation Learning}

\begin{figure}[]   
	
	\flushright
	
	\includegraphics[width=0.45\textwidth]{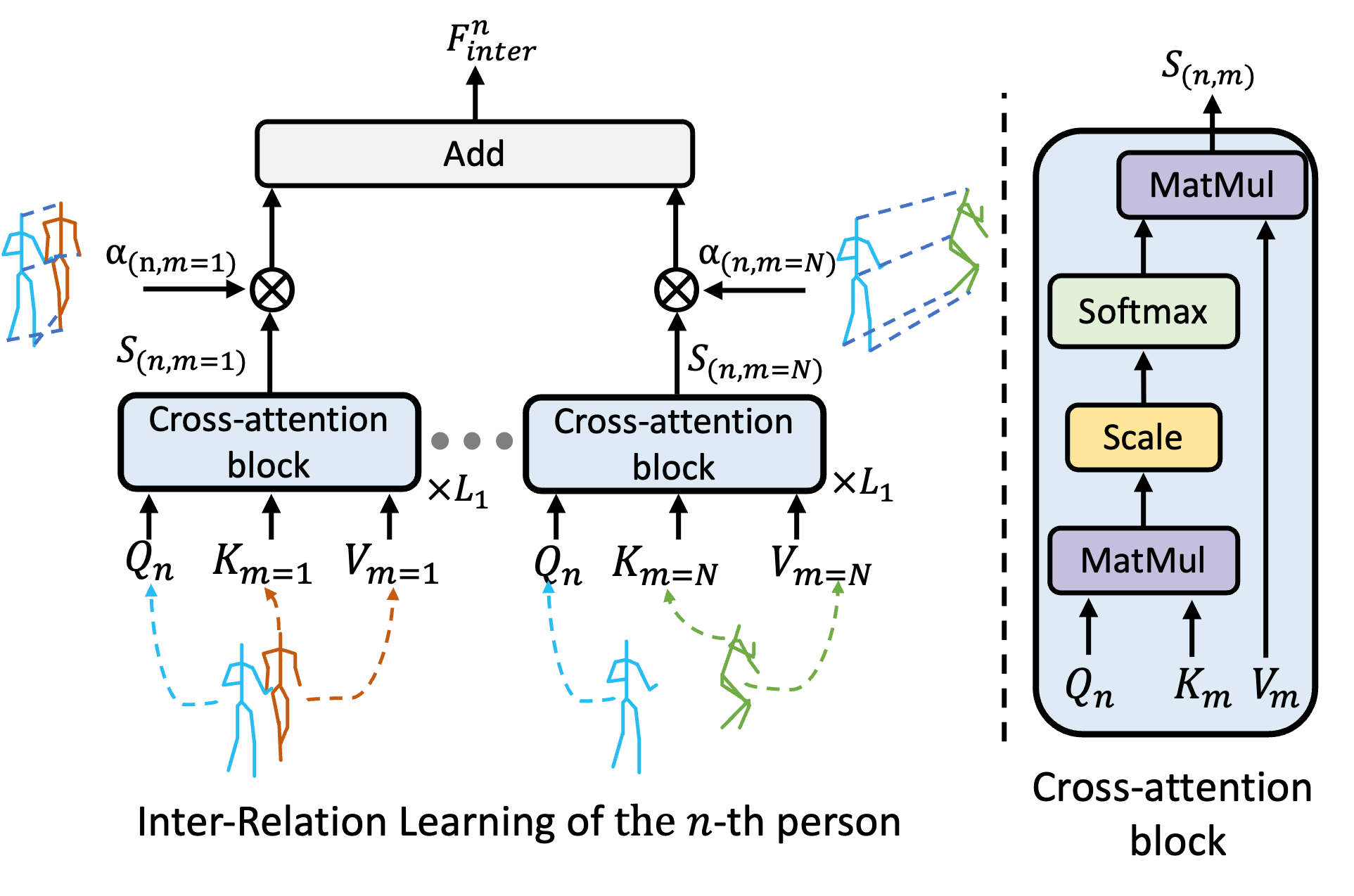}

	\caption{The illustration of inter-relation learning. The figure above represents the calculation of inter-relation feature of the $n$-th person. $n,m$ denote two distinct persons, person $n$ and person $m$. $\bigotimes $ denotes multiplication.}
	
	\label{arc2}
  
\end{figure}

\subsubsection{Inter-Relation Learning}

To explicitly learn the inter-relation $F_{inter}^{n}$ among multiple individuals, we introduce a novel module that utilizes cross-attention mechanism. The overall interaction score for each individual is obtained by weighted summing of the attention scores between that individual and the others, as shown in Fig. \ref{arc2}. The inter-relation feature for the $n$-th person  is calculated as follows:

\begin{equation}
F_{inter}^{n}=\sum_{m=1}^{N}\alpha _{\left ( n,m \right )}\times S_{\left (  n,m\right )}^{L_{1}}  \   \ (n\neq m)
\label{equation3}
\end{equation}
\noindent The $\alpha_{\left(n,m\right)}$ is the weight coefficient in the Eq. \ref{equation3}, and determined by the relative distance between the two individuals.  It is calculated as follows:

\begin{equation}
    \alpha_{\left(n,m\right)}=\frac{1}{\sigma  \left ( \lambda \right ) \times \mu  \left ( \sum_{t=1}^{T}x_{t}^{n}- x_{t}^{m}\right )+1}
\end{equation}
Where $\lambda$ is the learnable decay factor, and $\sigma \left (\cdot  \right )$ is the Sigmoid function, which constrains $\lambda$ within the range of 0 $\sim $ 1. $\mu$ denotes the average operation. The weight coefficient $\lambda$ decreases as the distance between the $n$-th person and the $m$-th person increases. 

The $S$ denotes the score of cross-attention mechanism. We use $l$ to indicate the number of cross-attention blocks, where $l\in  \left \{ 1,2,...,L_{1} \right \}$. The calculation can be expressed as:
\begin{equation}
\begin{aligned}
   S_{\left (  n,m\right )} ^{l}=\left\{\begin{matrix}
&\mathbf{ATT}{\left ( n,m \right )}\  \left ( l=1 \right )
\\ &\mathbf{ATT}{\left ( S_{\left (  n,m\right )} ^{l-1},m \right )}\  \left ( 1< l\leq L_{1} \right )

\end{matrix}\right. 
\end{aligned}
\end{equation}
$\mathbf{ATT}\left(\cdot\right)$ is the cross-attention mechanism.

\begin{equation}
    \mathbf{ATT}{\left ( n,m \right )} = MH\left ( softmax \left (\frac{Q^{n} \left (K^{m}  \right )^{T}}{\sqrt{d_{k^{m}}}}  \right )V^{m}\right )
\end{equation}
$MH(\cdot)$ denotes the computing of multi-head attention. $Q^{n}$ denotes the query of the $n$-th person , $V^{m}$ and $K^{m}$ denote the value and key of the $m$-th person. $\sqrt{d_{k^{m}}} $ is the feature size of the key $K^{m}$.

\subsubsection{Intra-Relation Learning}

\begin{figure}[t] 
	
	\centering
	
	\includegraphics[width=0.45\textwidth]{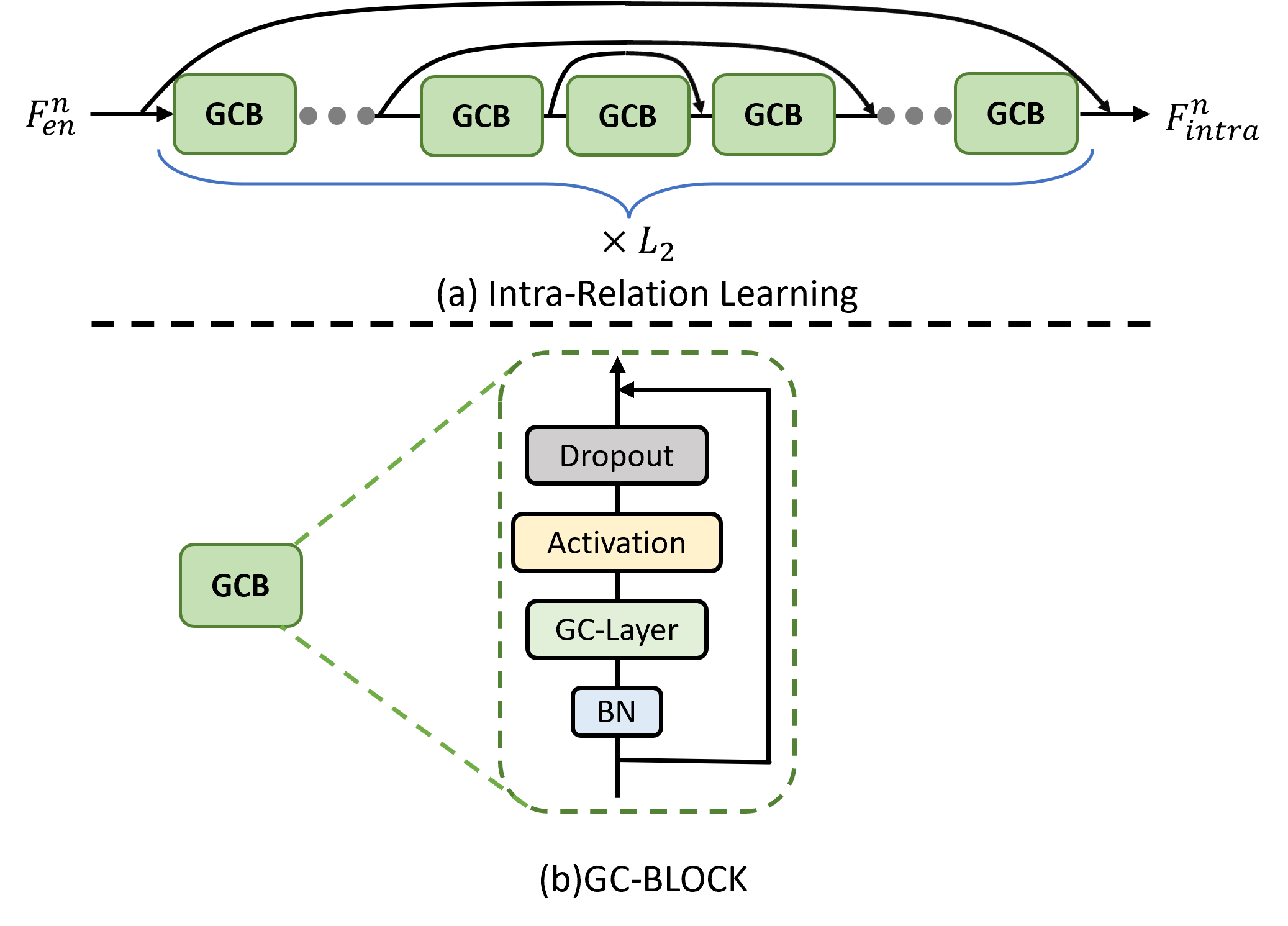}

	\caption{The illustration of intra-relation learning. (a)We utilize GC-blocks with symmetric residual connection to accomplish intra-relation learning. (b) The architecture of each GC-block. }
	
	\label{arc3}
  
\end{figure}

In a mean time, we utilize GCN to exploit individual’s intra-relations, as shown in Fig. \ref{arc3}. We feed $F_{en}^{n}$ into the GC-blocks to get the intra-relation information $F_{intra}^{n} = {H}^{n,L_{2}}$.

 \begin{equation}
 \begin{aligned}
H^{n,l}= \begin{cases}
 &Drop\left ( \sigma \left ( A^{l} \left ( BN\left ( F_{en}^{n}  \right ) \right )W^{l}\right ) \right )\ \ (l=1)\\ 
 &Drop\left ( \sigma \left ( A^{l} \left ( BN\left ( H^{n,l-1} \right ) \right )W^{l}\right ) \right )\ \ (1 <l\leqslant L_{2})
 \end{cases}
\end{aligned}
\end{equation}
 Where $l$ represents the $l$-th GC-block (in our method $1\leq l\leq L_{2}$), and $H^{n,l}$ represents the output of the $l$-th GC-block of the $n$-th person. $Drop(\cdot)$ is a regularization, $\sigma(\cdot)$ is an activation function, and the ${A}^{l}$ represents the adjacent matrix of the $l$-th GC-block. The matrix ${A^{l}}\in \mathbb{R}^{C\times C}$ is specifically designed to describe the relationships between nodes in GCN, where $C=J\times 3$ denotes the 3-dimension of $J$ input body joints. $BN(\cdot)$  represents Batch Normalization and $W^{l}$ is the trainable parameter matrix. Ultimately, we derive the intra-relation features $F_{intra}^{n}$ of the  $n$-th person.

\subsection{Relation Aggregation}

\begin{figure}[t]   
	
	\centering
	
	\includegraphics[width=0.45\textwidth]{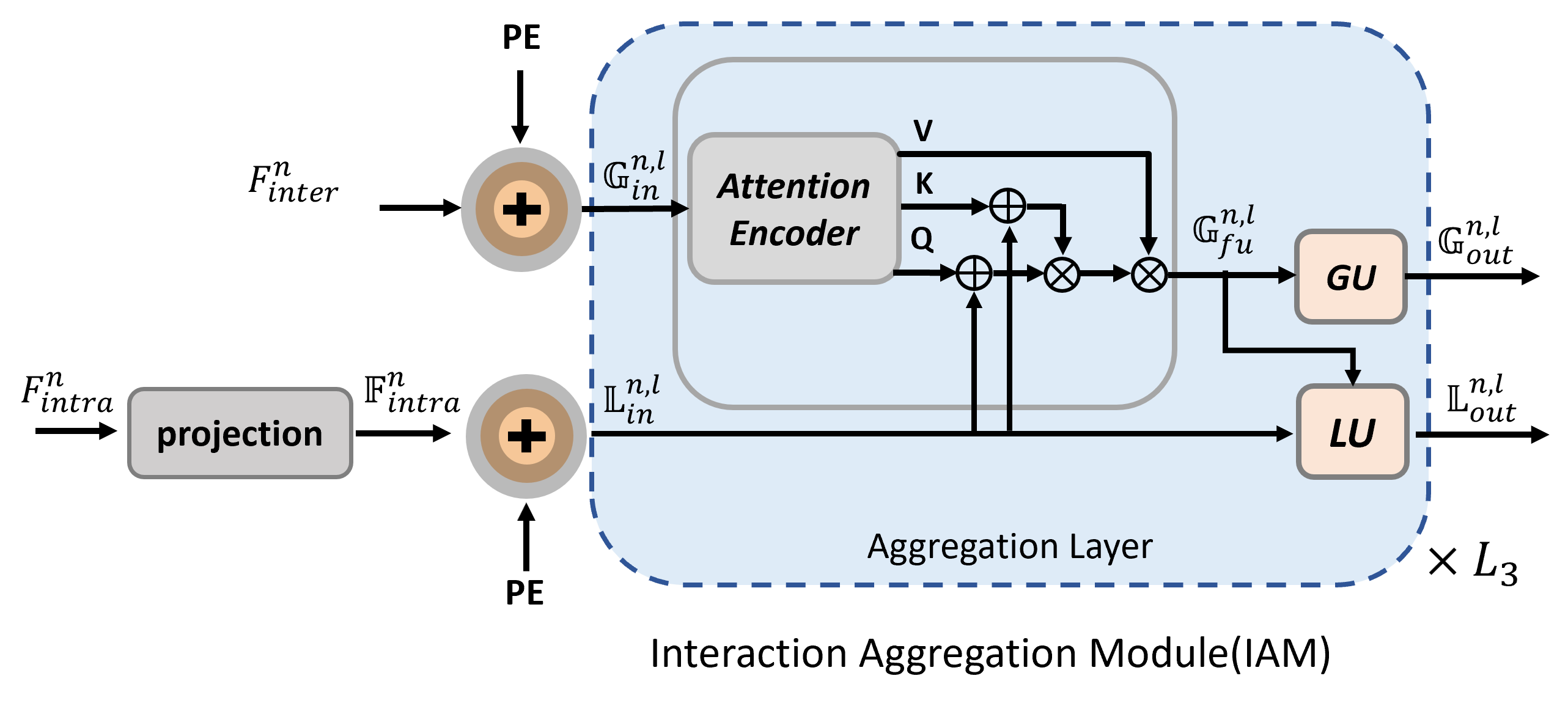}

	\caption{The computational pipeline of Interaction Aggregation Module (IAM). We introduce an aggregation-attention mechanism to integrate different features from two branches and update both features at each layer.} 
	
	\label{arc4}
  
\end{figure}

In this section, we introduce a novel feature aggregation module named IAM. It adopts a variant attention mechanism to aggregate intra\&inter-relation features, as illustrated in Fig. \ref{arc4}.

To enhance the integration of the two types of relations, we project the intra-relation features into a new dimension, which can be formulated as follows:

\begin{equation}
\begin{aligned}
\mathbb{F}_{intra}^{n} = &\mathbf{PROJ}\left ( F_{intra}^{n}\right )\\
= &\sigma \left ( F_{intra}^{n}\mathbf{W}_{1}\right )+\\
&\sigma\left ( \mathbf{W}_{2}\left(F_{intra}^{n}\mathbf{W}_{1}\right)\otimes \mathbf{W}_{3}\left(F_{intra}^{n}\mathbf{W}_{1}\right) \right ) 
\end{aligned}
\end{equation}
Where $\mathbf{W}_{1}$, $\mathbf{W}_{2}$ and $\mathbf{W}_{3} $ are learnable
parameters, $\otimes $ represents Hadamard product.

We then introduce a novel Interaction Aggregation Module (IAM) to enable our predictions to simultaneously utilize both inter\&intra-relation features, as shown in Fig. \ref{arc4}:
\begin{equation}
    \mathbb{L}_{in}^{n,l}=\left\{\begin{matrix}\mathbf{PE}\left(\mathbb{F}_{intra}^{n}\right)+\mathbb{F}_{intra}^{n}  \left (l=1 \right ) 
\\ \mathbb{L}_{out}^{n,l-1}  \left (1 < l\leq L_{3}  \right )

\end{matrix}\right.
\end{equation}

\begin{equation}
\mathbb{G}_{in}^{n,l}=\left\{\begin{matrix}\mathbf{PE}\left(F_{inter}^{n}\right)+F_{inter}^{n}\left (l=1 \right ) 
\\ \mathbb{G}_{out}^{n,l-1}\left (1< l\leq L_{3} \right )
\end{matrix}\right.
\end{equation}
Where ``$\mathbf{PE\left(\cdot\right)}$'' denotes the operation of position encoding. $\mathbb{L}_{in}^{n,l}$ and $\mathbb{G}_{in}^{n,l}$  denotes the input of $l$-th aggregation layer of the $n$-th person, and $\mathbb{L}_{out}^{n,l}$ and $\mathbb{G}_{out}^{n,l}$  denotes the output of $l$-th aggregation layer of the $n$-th person.
To explicitly incorporate the inter-relations and intra-relations, we design a novel aggregation layer with inter\&intra attention mechanism, formulated as:
{\small
\begin{equation}
\begin{aligned}
 \mathbb{G}_{fu}^{n,l}&=\mathcal{F} \left (  \mathbb{G}_{in}^{n,l}, \mathbb{L}_{in}^{n,l}\right )\ \ \ \ \ \ \ \ \ (1\leq  l\leq L_{3})\\  
        &=MHA(f_{QKV}(\mathbb{G}_{in}^{n,l}),\mathbb{L}_{in}^{n,l})\\
        &=MH\left ( softmax \left (\frac{\left ( Q^{\mathbb{G}}+\mathbb{L}_{in}^{n,l}\right )\left (K^{\mathbb{G}}+ \mathbb{L}_{in}^{n,l} \right )^{T}}{\sqrt{d_{k}}}  \right )V^{\mathbb{G}}\right ) \label{Synchronous Layer}
\end{aligned}
\end{equation} }
Where $MHA(\cdot)$ refers to a variant of multi-head attention. In Eq. \ref{Synchronous Layer}, we generate the Query/Key/Value for the $\mathbb{G}_{in}^{n,l}$ by the $f_{QKV}$. We create a special query by adding the embedded intra-relation information $\mathbb{L}_{in}^{n,l}$ to the initial Query derived from the inter-relation $Q^{\mathbb{G}}$. The resulting features $\mathbb{G}_{fu}^{n,l}$ effectively capture both inter\&intra-relations.
%Our strategy of aggregation is more comprehensive and explicit than JRT \cite{xu2023joint}, which involves computing the attention score of inter-relation before incorporating the intra-relation. 

Then we update our inter/intra-relations with a global update layer (GU) and a local update layer (LU):
\begin{align}
&\mathbb{G}_{out}^{n,l}= f_{GU}\left(LN\left (  \mathbb{G}_{fu}^{n,l}\right)\right )+\mathbb{G}_{in}^{n,l}\\
&  \mathbb{L}_{out}^{n,l}= f_{LU}\left(LN\left(\mathbb{G}_{syn}^{n,l},\mathbb{L}_{in}^{n,l}\right)\right )+\mathbb{L}_{in}^{n,l}
\end{align}
Where $f_{GU}\left(\cdot\right)$ is implemented by a MLP block, $f_{LU}\left(\cdot\right)$ is consist of a 1-layer Transformer. $LN(.)$ represents Layer Normalization \cite{ba2016layer}. 
\subsection{Decoder}
After obtaining the refined features $ \mathbb{G}_{out}^{n,L_{3}}$ and $ \mathbb{L}_{out}^{n,L_{3}}$, we concatenate these and feed them into the decoder:
\begin{equation}
    {\widehat{Y}^{n}} =\mathbf{FC}\left ( \mathbf{Concat}\left (\mathbb{G}_{out}^{n,L_{3}}, \mathbb{L}_{out}^{n,L_{3}} \right ) \right ) 
\end{equation}

\noindent Following this step, we can finally get the prediction ${\widehat{Y}^{n}}$ of the $n$-th person. 

\subsection{Loss functions}

We define the final loss function $\mathcal{L} $ as the sum of $\mathcal{L}_{p}$ and $\mathcal{L}_{v}$ to train the model jointly. The final loss function can be
expressed as follows:
\begin{equation}
\begin{aligned}
    \mathcal{L} = &\mathcal{L}_{p}^{2} + \mathcal{L}_{v}^{2}\\
    =&\left [ \ell_{2}\left (\widehat{Y}^{n},  Y^{n} \right ) \right ]^{2}+\left [ \ell_{2}\left (\widehat{Y}_{v}^{n},  Y_{v}^{n} \right ) \right]^{2}
\end{aligned}
\end{equation}
$\mathcal{L}_{p}$ aims to minimize the $\ell_{2}$ -norm between the predicted
motion $\widehat{Y}^{n}$ and ground-truth $Y^{n}$. $\mathcal{L}_{v}$ aims to minimize the $\ell_{2}$ -norm between the predicted
motion $\widehat{Y}_{v}^{n}$ and ground-truth of velocity ${Y}_{v}^{n}$ calculated by position difference.

\section{Experiment and discussions}
% \usepackage{multirow}

\subsection{Baselines}
We select a range of base models to verify the effectiveness of our method, including classical approaches such as LTD \cite{mao2019learning}, TRiPOD \cite{9709907}, DViTA \cite{parsaeifard2021learning}, and MRT \cite{wang2021multi}, the single-person motion prediction method TCD \cite{saadatnejad2023generic}, as well as state-of-the-art models JRT \cite{xu2023joint} and TBIFormer \cite{peng2023trajectory} as our baselines. LTD \cite{mao2019learning} is a foundational model in the field of human motion prediction that encodes temporal information by operating in trajectory space. TRiPOD \cite{9709907} first models interactions in human-to-human and human-to-object. DViTA \cite{parsaeifard2021learning} divides multi-person prediction tasks into several single-person prediction tasks. MRT \cite{wang2021multi} emphasizes the importance of interaction in multi-person prediction tasks by adopting a multi-range Transformer. JRT \cite{xu2023joint} explores the physical relations of the human body in conjunction with interactions to yield noteworthy outcomes. TBIFormer \cite{peng2023trajectory} addresses the human-to-human interactions by focusing on skeletal body parts. It converts all pose sequences into Multi-Person BodyPart sequences, utilizing these transformed sequences to learn the dynamics of body parts for both inter- and intra-individual interactions.

\begin{table*}[t]
\centering
\caption{Experimental results in VIM on 3DPW and 3DPW-RC test sets. The best results are highlighted in bold. "-" indicates that the data are not given in the paper and cannot be reproduced without code. "*" indicates that the data are not given in the paper and we reproduced with the paper's code.}

\begin{tabular}{cl|cccccc|cccccc}
\hline
\multicolumn{2}{c|}{\multirow{2}{*}{Method}} & \multicolumn{6}{c|}{\begin{tabular}[c]{@{}c@{}}3DPW\\ (2 persons)\end{tabular}}                                                              & \multicolumn{6}{c}{\begin{tabular}[c]{@{}c@{}}3DPW-RC\\ (2 persons)\end{tabular}}                                                            \\ \cline{3-14} 
\multicolumn{2}{c|}{}                        & AVG           & 100          & 240           & 500           & 640           & 900           & AVG           & 100          & 240           & 500           & 640           & 900           \\ \hline
\multicolumn{2}{c|}{LTD \cite{mao2019learning}}                     & 76.7          & 22.0         & 41.1          & 81.0          & 100.2         & 139.7         & 62.2          & 20.2         & 37.2          & 68.6          & 81.3          & 103.5         \\
\multicolumn{2}{c|}{TRiPOD \cite{9709907}}                   & 84.2          & 31.0         & 50.8          & 84.7          & 104.1          & 150.4         & -         & -        & -         & -         & -         & -        \\
\multicolumn{2}{c|}{DViTA \cite{parsaeifard2021learning}}                   & 65.7          & 19.5         & 36.9          & 68.3          & 85.5          & 118.2         & 57.4          & 15.0         & 32.3          & 62.7          & 76.3          & 100.6         \\

\multicolumn{2}{c|}{MRT \cite{wang2021multi}}                     & 59.2          & 21.8         & 39.1          & 65.1          & 75.9          & 94.1          & 52.3          & 20.8         & 36.4          & 58.2          & 66.6          & 79.4          \\

\multicolumn{2}{c|}{TCD* \cite{saadatnejad2023generic}}                     & 51.0          & 10.8         & 24.8          & 55.1          & 68.3          & 96.1          & 43.3          & 10.2         & 26.3          & 50.1          & 57.8          & 72.1          \\
\multicolumn{2}{c|}{TBIFormer* \cite{peng2023trajectory}}                     & 48.4         & 9.8          & 22.5          & 50.6          & 63.1          & 94.2         & 41.5          & 9.7        & 24.0          & 47.7          & 55.3          & 70.8         
\\
\multicolumn{2}{c|}{JRT \cite{xu2023joint}}                     & 47.2          & 9.5          & 22.1          & 48.7          & 62.8          & 92.8          & 39.5          & 9.5          & 21.7          & 44.1          & 53.4          & 68.8          
\\

\hline

\multicolumn{2}{c|}{Ours}          & \textbf{46.5}          & \textbf{9.3}         & \textbf{21.5}          & \textbf{47.3} & \textbf{62.1}          & \textbf{92.3}          & \textbf{39.0} & \textbf{9.2} & \textbf{21.6} & \textbf{43.6} & \textbf{52.6} & \textbf{68.5} \\ \hline
\label{3dpw-vim}
\end{tabular}
\end{table*}
\begin{table}[ht]
\centering
\caption{Experimental results in MPJPE on CMU-Mocap (left) and
MuPoTS-3D (right) test sets. The best results are highlighted in
bold. "*" indicates that the data are not given in the paper and we reproduced with the paper's code.}

\begin{tabular}{llccc|ccc}
\hline
\multicolumn{2}{c|}{\multirow{2}{*}{Methods}}    & \multicolumn{3}{c|}{\begin{tabular}[c]{@{}c@{}}CMU-Mocap\\ (3 persons)\end{tabular}} & \multicolumn{3}{c}{\begin{tabular}[c]{@{}c@{}}MuPoTS-3D\\ (2$\sim$3 persons)\end{tabular}} \\ \cline{3-8} 
\multicolumn{2}{c|}{}                            & 1s     & 2s     & 3s     & 1s     & 2s     & 3s    \\ \hline
\multicolumn{2}{c|}{LTD \cite{mao2019learning}}                        & 13.7     & 21.9     & 32.6     & 11.9     & 18.1     & 23.4    \\
\multicolumn{2}{c|}{HRI \cite{mao2020history}}                        & 14.9     & 26.0     & 30.7     & 9.4      & 16.8     & 22.9    \\
\multicolumn{2}{c|}{TCD* \cite{saadatnejad2023generic}}                        & 10.2      & 16.1     & 19.5     & 9.0     & 15.8     & 21.7    \\
\multicolumn{2}{c|}{MRT \cite{wang2021multi}}                        & 9.6      & 15.7     & 21.8     & 8.9     & 15.9     & 22.2    \\

\multicolumn{2}{c|}{TBIFormer* \cite{peng2023trajectory}} & 8.0     & 13.4     & 19.0     & 8.7      & 15.1     & 20.9  
\\

\multicolumn{2}{c|}{JRT \cite{xu2023joint}} & 8.3      & 13.9     & 18.5     & 8.9      & 15.5     & 21.3  
\\
\hline
\multicolumn{2}{c|}{Ours}                       &    \textbf{7.8}      &  \textbf{12.8}        &  \textbf{16.9}        &   \textbf{8.5}    &  \textbf{14.7}        &  \textbf{20.1}       \\ \hline
\end{tabular}
\label{cmu/mopots}
\end{table}

\begin{table}[ht]
\centering
\caption{Experimental results in MPJPE on Mix1 (left) and
Mix2 (right) test sets. The best results are highlighted in
bold. "*" indicates that the data are not given in the paper and we reproduced with the paper's code.}

\begin{tabular}{llccc|ccc}
\hline
\multicolumn{2}{c|}{\multirow{2}{*}{Methods}}    & \multicolumn{3}{c|}{\begin{tabular}[c]{@{}c@{}}Mix1\\ (9$\sim$15 persons)\end{tabular}} & \multicolumn{3}{c}{\begin{tabular}[c]{@{}c@{}}Mix2\\ (11 persons)\end{tabular}} \\ \cline{3-8} 
\multicolumn{2}{c|}{}                            & 1s     & 2s     & 3s     & 1s     & 2s     & 3s    \\ \hline
\multicolumn{2}{c|}{LTD \cite{mao2019learning}}                        & 21.0     & 31.9     & 41.5     & 17.2     & 25.8     & 34.5    \\
\multicolumn{2}{c|}{HRI \cite{mao2020history}}                        & 18.0     & 31.4     & 42.1     & 16.0      & 27.1     & 36.7    \\
\multicolumn{2}{c|}{MRT \cite{wang2021multi}}                        & 17.3      & 29.9     & 39.7     & 12.9     & 20.9     & 28.2    \\

\multicolumn{2}{c|}{TBIFormer* \cite{peng2023trajectory}} & 14.0      & 24.4     & 31.2     & 12.3      & 18.2     & 26.1  
\\
\hline
\multicolumn{2}{c|}{Ours}                       &    \textbf{13.7}      &  \textbf{23.7}        &  \textbf{29.7}        &   \textbf{11.8}    &  \textbf{17.5}        &  \textbf{25.2}       \\ \hline
\end{tabular}

\label{mix1/mix2}
\end{table}

\subsection{Datasets}
We adopt four multi-person motion prediction benchmarks
in the experiments: 3DPW \cite{Marcard_2018_ECCV}, 3DPW-RC \cite{xu2023joint}, CMU-Mocap \cite{cmumocap2003},
MuPoTs-3D \cite{mehta2018single} and synthesized datasets Mix1\&Mix2 \cite{wang2021multi}. Details of the four datasets are presented below.

\textbf{3DPW \cite{Marcard_2018_ECCV}}
3D Poses in the Wild Dataset (3DPW) is a large-scale
3D motion dataset collected by moving mobile phone
cameras. It contains 60 videos and about 68,000 frames
covering multiple scenarios and actions. In this
paper, we follow the setting of \cite{adeli2020socially,9709907,10194334,xu2023joint}. Each scene contains two persons, and our goal is to predict future 900ms (14 frames) motion using the historical
1030ms (16 frames) motion.

\textbf{3DPW-RC \cite{xu2023joint}} According to the findings from the Joint-Relation Transformer \cite{xu2023joint}, camera movement in the 3DPW dataset induces a significant unnatural drift, negatively impacting the modeling of multi-person interactions. The 3DPW-RC dataset subtracts the estimated camera velocity for better evaluation. 

\textbf{CMU-Mocap \cite{cmumocap2003}} The Carnegie Mellon University Motion Capture Database (CMU-Mocap) consists of data from 112 subjects. Most scenes capture the movements of one person, and only 9 scenes include the movements and interactions of two persons. The study in \cite{wang2021multi} combines samples from both one-person and two-person scenes to create sequences featuring three individuals. We use the training and test sets provided by Wang et al. \cite{wang2021multi}
to train and evaluate our model. We aim to predict future 3000ms (45 frames) motion using the historical 1000ms (15 frames) motion.

\textbf{MuPoTS-3D \cite{mehta2018single}} Multiperson Pose Test Set in 3D (MuPoTS-3D) consists of over 8000 frames collected from 20 sequences with 8 subjects. Following previous
works \cite{wang2021multi,xu2023joint}, we evaluate our model’s performance with
the same segment length as CMU-Mocap on the test set.

\textbf{Mix1\&Mix2 \cite{wang2021multi}} In order to evaluate the performance of
our proposed model in scenarios involving a larger number of
individuals, we adopt the methodology presented in the MRT \cite{wang2021multi} paper. We sample data from the CMU-Mocap and Panoptic \cite{joo2016panoptic} datasets to generate the Mix1 traning set. This
training set contains approximately 3,000 samples, each featuring
9 to 15 people in the scene.
Next we combine CMU-Mocap, MuPoTS-3D and 3DPW data, namely Mix2. There are 11 persons in each scene of its 400 samples.

\subsection{Implementation Details}
In practice, for the 3DPW and 3DPW-RC datasets, we set the input length $T = 16$, the output
length $P = 14$, and the number of person $N=2$. For the CMU-Mocap and MuPoTS-3D datasets, we set $T = 15$, $P = 45$ and $N=3$.

Our model consists of $L_{1}=4$ cross-attention blocks with $H=8$ attention heads for inter-relation learning, $L_{2}=13$ GC-blocks for intra-relation learning, and $L_{3}=4$ aggregation layers in the IAM. We pre-train the model on the AMASS \cite{DBLP:journals/corr/abs-1904-03278} dataset following
previous works \cite{xu2023joint,vendrow2022somoformer}, which provides massive motion sequences.
We utilize the PyTorch deep learning framework to develop
our models and optimize the training with AdamW \cite{loshchilov2017decoupled} optimizer. The
learning rate is set to $1\times 10^{-5}$ for both pre-train and finetune and decay by 0.8 every
10 epochs. The batch size is set to 256 for pre-train, 128 for finetune. The training is performed on an
NVIDIA 3080Ti GPU for 100 epochs.

\subsection{Metrics}
\textbf{VIM \cite{9709907}} We adopt the Visibility-Ignored Metric (VIM) to measure the displacement in the joint vector, which has a dimension of $J\times3$, following previous works. This metric is used on the 3DPW and 3DPW-RC datasets.
\begin{equation}
   \mathbf{VIM@t}=\frac{1}{N}\sum_{n=1}^{N}\sqrt{\sum_{j=1}^{J}(Y_{nj}^{t}-\widehat{Y}_{nj}^{t})}
\end{equation}

\textbf{MPJPE \cite{6682899}} Mean Per Joint Position Error (MPJPE) is a commonly used metric in human motion prediction, which calculates the average Euclidean distance between the prediction and the ground truth
of all joints. We use this metric on CMU-Mocap and MuPoTS-3D.

\begin{equation}
    \mathbf{MPJPE}=\frac{1}{P}\frac{1}{N}\frac{1}{J}\sum_{t=1}^{p}\sum_{n=1}^{N}\sum_{j=1}^{J}\left \| Y_{nj}^{t}- \widehat{Y}_{nj}^{t}\right \|_{2}
\end{equation}

\subsection{Quantitative Results}

\subsubsection{Results on 3DPW and 3DPW-RC} For a fair comparison, we follow the same VIM criterion as established in previous works \cite{parsaeifard2021learning,xu2023joint,wang2021simple,wang2021multi,9709907}. The experimental results in VIM on the 3DPW and 3DPW-RC datasets are shown in Table.\ref{3dpw-vim}, where our method achieves the best performance on each dataset. Compared
to the current state-of-the-art method Joint-Relation Transformer \cite{xu2023joint}, our method reduces the VIM on AVG from 47.2 to 46.5 on 3DPW dataset and from 39.5 to 39.0 on 3DPW-RC dataset. These improvements demonstrate the effectiveness of our method. Traditional single-person motion prediction methods DViTA \cite{parsaeifard2021learning} and LTD \cite{mao2019learning} focuses on trajectory and ignores human interaction in multi-person motion prediction, so their experimental results are uncompetitive compared to subsequent methods. TRiPOD \cite{9709907} employs GAT 
\cite{veličković2018graph} to capture the connections between different targets. However, GAT is limited to capturing local information, as it performs attention operations only on adjacent nodes, thus failing to capture global information. MRT \cite{wang2021multi} utilizes a multi-range Transformer to model interactions and human trajectories separately, it employs a Transformer decoder instead of designing a module to aggregate two types of information. Since TCD \cite{saadatnejad2023generic} is designed for single-person  motion prediction, it lacks the capability to capture human-to-human interactions. The results highlight the significance of modeling interactions in multi-person motion prediction.

\begin{figure}[t]

\begin{minipage}{1\linewidth}

\centering
\caption{Experimental results in VIM on 3DPW test sets. We verify the effectiveness of IAM incorporating different base models, TRiPOD \cite{9709907}, MRT \cite{wang2021multi}, and JRT\cite{xu2023joint}.The best results are highlighted in bold. "+" denotes base models enhance with our IAM. "$\oplus$" indicates that the fusion module is replaced with our IAM.}
{\scriptsize
\begin{tabular}{c|cccccc}
\hline
\multirow{2}{*}{Method} & \multicolumn{6}{c}{3DPW}                  \\ \cline{2-7} 
                        & AVG  & 100  & 240  & 500  & 640   & 900   \\ \hline
TRiPOD \cite{9709907}                  & 84.2 & 31.0 & 50.8 & 84.7 & 104.1 & 150.4 \\
TRiPOD+IAM              & \textbf{77.9} & \textbf{28.6} & \textbf{46.8} & \textbf{79.2} & \textbf{95.1}  & \textbf{140.2} \\
MRT \cite{wang2021multi}                    & 59.2 & 21.8 & 39.1 & 65.1 & 75.9  & 94.1  \\
MRT+IAM                 & \textbf{54.3} & \textbf{18.1} & \textbf{32.3} & \textbf{59.9} & \textbf{68.2}  & \textbf{93.2}  \\ 
JRT \cite{xu2023joint}                    & 47.2 & 9.5 & 22.1 & 48.7 & 62.8  & 92.8  \\
JRT$\oplus $IAM                 & \textbf{46.8} & \textbf{9.5} & \textbf{21.8} & \textbf{48.1} & \textbf{62.5}  & \textbf{92.8}  \\ \hline
Ours (w/o IAM)                  & 48.0 & 9.5 & 22.4 & 49.1 & 64.8  & 94.5  \\ 
Ours               & \textbf{46.5} & \textbf{9.3}  & \textbf{21.5} & \textbf{47.3} & \textbf{62.1}  & \textbf{92.3}  \\ \hline
\end{tabular}
}

\label{IAM_tab}
\end{minipage}

\begin{minipage}{1\linewidth}
        \centering
    	\includegraphics[width=1\textwidth]{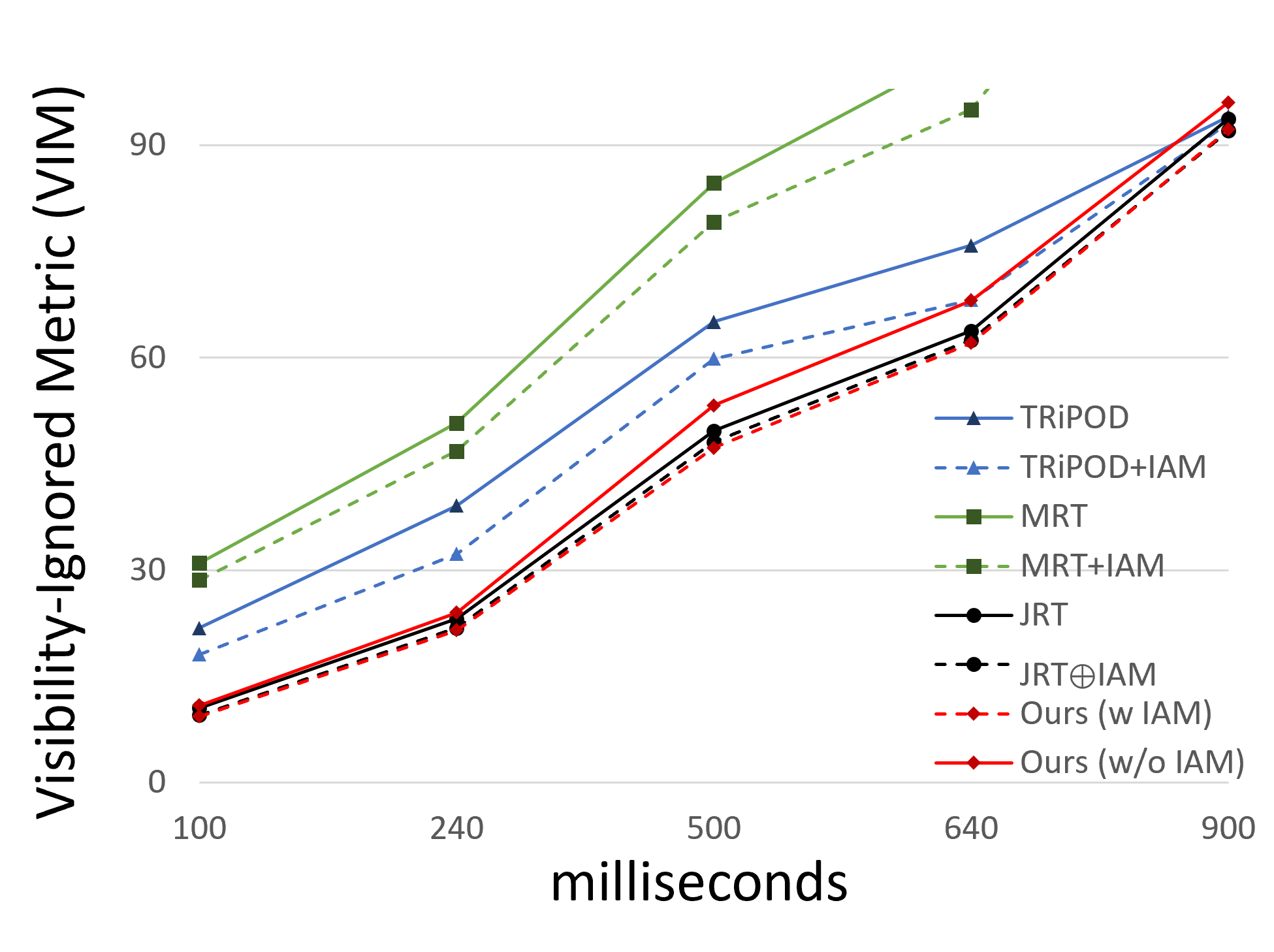}
		\caption{The line chart of experimental results on the 3DPW \cite{Marcard_2018_ECCV} dataset. The green line denotes baseline model MRT \cite{wang2021multi}. The blue line denotes baseline model TRiPOD \cite{9709907}. The purple line denotes baseline model JRT \cite{xu2023joint}. The red line denotes our method without IAM. The green, blue, and purple dashed lines denote the baseline models enhanced with our IAM. The red dashed line denotes our method with IAM. Obviously, the utilization of our IAM yields a more accurate prediction.}
		\label{IAM_fig}
\end{minipage}
\end{figure}

\begin{figure}[ht] 
	
	\centering
	
	\includegraphics[width=0.4\textwidth]{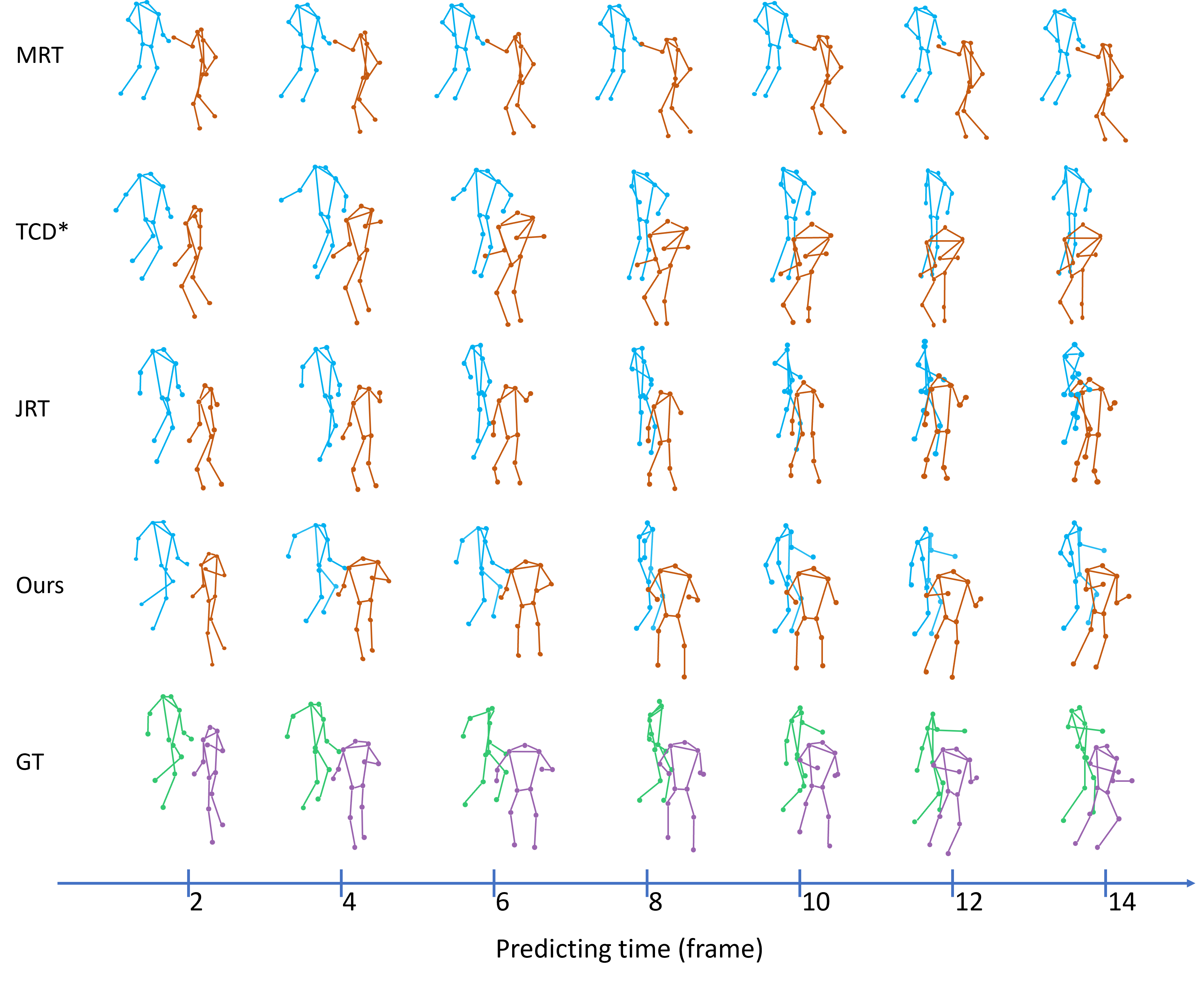}

	\caption{Visualization comparison on 3DPW-RC dataset.
We compare the prediction by our method and three previous methods.
Our method generates a more precise motion prediction.}
	\label{visual1}

\end{figure}

\subsubsection{Results on CMUMocap, MuPoTS3D and Mix1\&Mix2} We also compare the results in MPJPE on the CMU-Mocap and MuPots-3D datasets between our method and several other approaches, including MRT \cite{wang2021multi} and Joint-Relation Transformer \cite{xu2023joint}, as well as two single-person motion prediction methods: HRI \cite{mao2020history} and LTD \cite{mao2019learning}. We provide 15 observed frames (1s) 
as input to predict the subsequent 45 frames (3s), and report
the MPJPE at 1, 2, and 3 seconds into the future, as shown in Table. \ref{cmu/mopots}. Our method achieves state-of-the-art results on CMU-Mocap and MuPots-3D, demonstrating its accuracy and strong generalization capabilities. In addition, we conduct experiment on the Mix1\&Mix2 datasets to validate our method's ability to handle sophisticated scenes involving more than 9 persons, as shown in Table. \ref{mix1/mix2}. The results demonstrate that our method excels at managing more complex situations.

\begin{figure*}[ht]   
	
	\centering
	
	\includegraphics[width=0.8\textwidth]{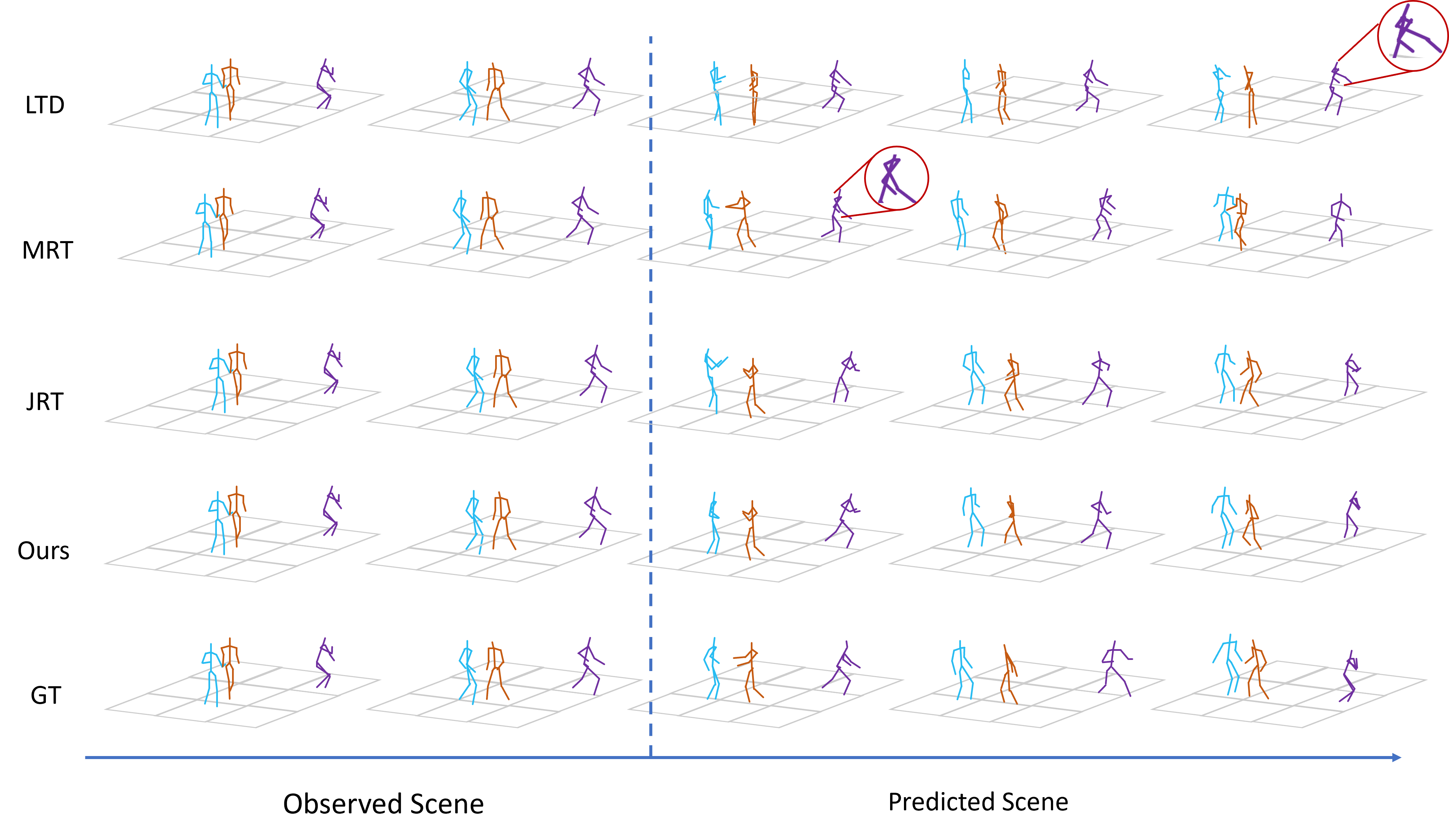}

	\caption{Visualization comparison on CMU-Mocap dataset.
We compare the prediction by our method and three previous methods.
Our method generates a more natural and accurate motion prediction.}
	\label{visual2}

\end{figure*}

\begin{figure*}[ht] 
	
	\centering
	
	\includegraphics[width=1\textwidth]{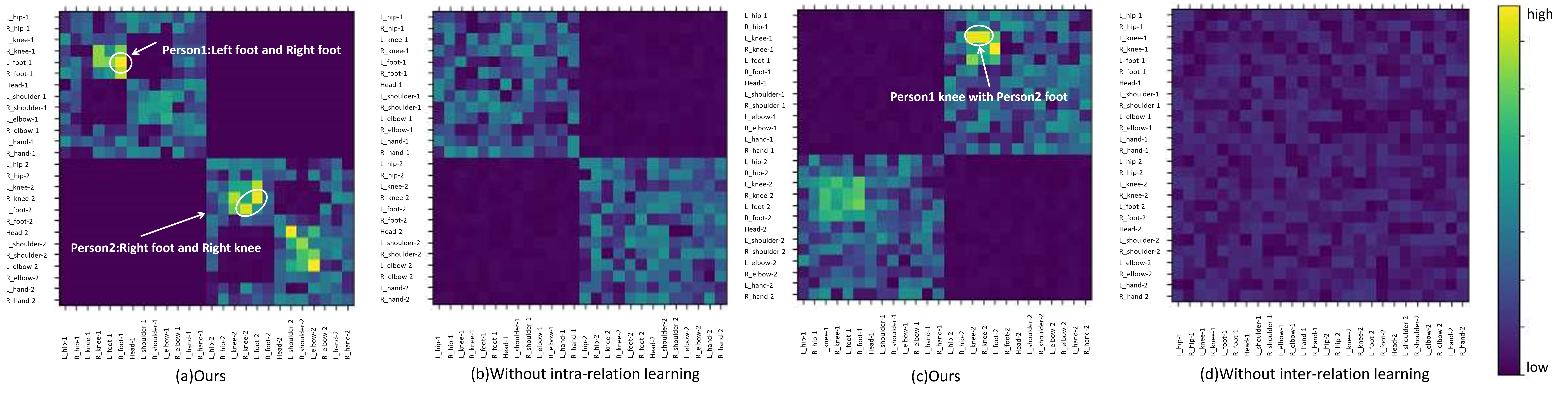}

	\caption{Attention visualization of intra\&inter-relations on 3DPW-RC dataset. We compare our method and baseline without intra-relation or inter-relation modeling. (a)Our method which adopts intra-relation learning. (b)Baseline without intra-relation learning. Lighter colors indicate higher attention scores and higher connections. (c)Our method which adopts inter-relation learning. (d)Baseline without inter-relation learning. Lighter colors indicate higher attention scores and higher connections.}
	\label{attention visual}
 \end{figure*}

\subsubsection{Effectiveness of IAM}
In order to verify the effectiveness of IAM, we designed
an study on the 3DPW test set using DViTA \cite{9709907}, MRT \cite{wang2021multi}, and JRT\cite{xu2023joint} as the base models. The experimental results are shown in Table. \ref{IAM_tab} and Fig. \ref{IAM_fig}.
In these presentations, ``TRiPOD+IAM" and ``MRT+IAM" refer to the DViTA \cite{9709907}, MRT \cite{wang2021multi} enhanced with our IAM module. In addition, we replace the fusion module of JRT\cite{xu2023joint} with our IAM, denotes as ``JRT$\oplus $IAM". The results indicate that our IAM provides improvements to the base models, reducing the average error from 84.2 to 77.9 for TRiPOD, from 59.2 to 54.3 for MRT, and 47.2 to 46.8 for JRT. The results confirm that IAM effectively aggregates features from two different branches. These outcomes verify the strong ability of plug-and-play. Additionally, an ablation study was conducted to assess the impact of removing IAM from our method and the results demonstrate the importance of IAM.

% \usepackage{multirow}
% \usepackage{multirow}

\subsection{Qualitative Results}

\subsubsection{Visualization of prediction result}
We provide a qualitative
comparison on 3DPW-RC test set between our method and other recent
methods, including MRT \cite{wang2021multi}, FutureMotion \cite{wang2021simple} and JRT \cite{xu2023joint}, as shown in Fig. \ref{visual1}. Compared with the prediction of FutureMotion \cite{wang2021simple}, MRT \cite{wang2021multi}, and JRT \cite{xu2023joint}, our results are more natural and closer to the ground truth, particularly in the movement of the lower limbs. We also provide the visualization results on CMU-Mocap dataset to verify our method's effectiveness on scenes of 3 persons as shown in Fig. \ref{visual2}. We can notice that both LTD \cite{mao2019learning} and MRT \cite{wang2021multi} generate unnatural arm distortions that do not appear in our approach, as shown in the red circles. The visualization results on different datasets demonstrate the generalization and accuracy of our method.

\subsubsection{Visualization of attention scores}
We visualize the learned attention matrices in the first layer to demonstrate the effectiveness of the intra-relation learning and inter-interaction learning, as shown in Fig. \ref{attention visual}. We observe that in Fig. \ref{attention visual}(a) the values between" left foot and right foot" for person 1 and "right foot and knee" for person 2 are significantly higher compared to those in Fig. \ref{attention visual}(b), which lacks the intra-relation learning (The lighter the color, the higher the value). This provides evidence that intra-relation modeling is more conducive to highlighting relations within individual's joints. In Fig. \ref{attention visual}(c) and (d), we find that the inter-relation learning significantly differentiates the attention scores among different individuals. This is noticeable in the joints that exhibit large movements when a person performs an action, such as boxing, playing basketball.

\begin{table*}[h] 
\centering
\caption{Computational complexity analysis on 3DPW-RC dataset based on vim.}
% \usepackage{multirow}

% \usepackage{multirow}

\begin{tabular}{c|c|cl|cccccc}
\hline
\multirow{2}{*}{\begin{tabular}[c]{@{}c@{}}Number of Layers\\ $L_{i}$/$L_{e}$/$L_{s}$\end{tabular}} & \multirow{2}{*}{\begin{tabular}[c]{@{}c@{}}Parameters\\ (M)\end{tabular}} & \multicolumn{2}{c|}{\multirow{2}{*}{GFLOPs}} & \multicolumn{6}{c}{3DPW-RC}                                                                                                             \\ \cline{5-10} 
                                                                                                    &                                                                           & \multicolumn{2}{c|}{}                       & AVG                  & 100                  & 240                  & 500                  & 640                  & 900                  \\ \hline

\multicolumn{1}{c|}{TBIFormer\cite{peng2023trajectory}}                                                                               & \multicolumn{1}{c|}{6.1M}                                                     & \multicolumn{2}{c|}{1.7}                       & \multicolumn{1}{c}{41.5} & \multicolumn{1}{c}{10.2} & \multicolumn{1}{c}{24.0} & \multicolumn{1}{c}{47.7} & \multicolumn{1}{c}{55.3} & \multicolumn{1}{c}{71.4} \\
\multicolumn{1}{c|}{JRT\cite{xu2023joint}}                                                                               & \multicolumn{1}{c|}{6.7M}                                                     & \multicolumn{2}{c|}{1.2}                       & \multicolumn{1}{c}{39.5} & \multicolumn{1}{c}{9.5} & \multicolumn{1}{c}{21.7} & \multicolumn{1}{c}{44.1} & \multicolumn{1}{c}{53.4} & \multicolumn{1}{c}{68.8} \\\hline
Ours $L_{1}=1$/$L_{2}=3$/$L_{3}=1$                                                                       & 1.8M                                                                      & \multicolumn{2}{c|}{0.5}                    & 43.2                 & 10.9                 & 24.5                 & 48.2                 & 57.7                 & 75.1                 \\
Ours $L_{1}=2$/$L_{2}=3$/$L_{3}=2$                                                                       & 2.4M                                                                      & \multicolumn{2}{c|}{0.7}                    & 41.5                 & 10.5                 & 23.6                 & 45.5                 & 55.8                 & 72.3                 \\
Ours $L_{1}=4$/$L_{2}=6$/$L_{3}=4$                                                                       & 4.2M                                                                      & \multicolumn{2}{c|}{1.0}                    & 39.4                 & 9.3                 & 21.9                 & 44.2                 & 53.8                 & 69.1                 \\
Ours $L_{1}=8$/$L_{2}=13$/$L_{3}=8$                                                                      & 8.4M                                                                      & \multicolumn{2}{c|}{2.1}                    & 41.2                 & 10.6                 & 23.3                 & 45.7                 & 55.6                 & 70.9                 \\
Ours $L_{1}=8$/$L_{2}=26$/$L_{3}=8$                                                                      & 12.1M                                                                     & \multicolumn{2}{c|}{2.6}                    & 42.3                 & 10.9                 & 24.0                 & 47.0                 & 57.0                 & 72.4                 \\ \hline
Ours $L_{1}=4$/$L_{2}=13$/$L_{3}=4$                                                               & 5.1M                                                                      & \multicolumn{2}{c|}{1.1}                    & \textbf{39.1}        & \textbf{9.2}         & \textbf{21.6}        & \textbf{43.6}        & \textbf{52.6}        & \textbf{68.5}        \\ \hline
\end{tabular}

\label{paramater}
\end{table*}

\subsection{Ablation Study}

In this section, we present the model's complexity and the number of parameters. We also conduct the ablation study on the framework without velocity augment, without intra-relation learning, and without inter-relation learning. The results are shown in Table. \ref{paramater}, Table. \ref{velocity}, and Table. \ref{ablation on module}.
% \usepackage{multirow}

\begin{table}
\centering
\caption{Ablation study of velocity augment on 3DPW-RC based on VIM.}

\begin{tabular}{cl|cccccc}

\hline
\multicolumn{2}{c|}{\multirow{2}{*}{Ablation}} & \multicolumn{6}{c}{3DPW-RC}                                                                  \\ \cline{3-8} 
\multicolumn{2}{c|}{}                          & AVG           & 100          & 240           & 500           & 640           & 900           \\ \hline
\multicolumn{2}{c|}{w/o velocity}              & 39.6          & 9.8          & 22.1          & 44.7          & 53.6          & 69.0          \\
\multicolumn{2}{c|}{w velocity (Ours)}                      & \textbf{39.1} & \textbf{9.2} & \textbf{21.6} & \textbf{43.6} & \textbf{52.6} & \textbf{68.5} \\ \hline
\end{tabular}

\label{velocity}
\end{table}

\begin{table}[ht]
\centering
\caption{Ablation study of intra/inter-relation modeling on 3DPW-RC based on VIM.}

\begin{tabular}{cl|cccccc}

\hline
\multicolumn{2}{c|}{\multirow{2}{*}{Ablation}} & \multicolumn{6}{c}{3DPW-RC}                                                                  \\ \cline{3-8} 
\multicolumn{2}{c|}{}                          & AVG           & 100          & 240           & 500           & 640           & 900           \\ \hline
\multicolumn{2}{c|}{w/o intra}              & 41.6         & 10.6          & 23.3          & 45.7          & 55.6          & 70.9          \\
\multicolumn{2}{c|}{w/o inter}              & 42.4         & 10.9         & 24.0          & 47.0          & 57.0          & 72.4         \\
\multicolumn{2}{c|}{w/o intra\&inter}              & 46.7         & 12.5         & 28.6          & 53.5          & 62.8          & 76.3         \\
\multicolumn{2}{c|}{Ours}                      & \textbf{39.1} & \textbf{9.2} & \textbf{21.6} & \textbf{43.6} & \textbf{52.6} & \textbf{68.5} \\ \hline
\end{tabular}

\label{ablation on module}
\end{table}

\begin{table}[ht]
\centering
\caption{Ablation study of loss function.}

\begin{tabular}{cl|cccccc}

\hline
\multicolumn{2}{c|}{\multirow{2}{*}{Ablation}} & \multicolumn{6}{c}{3DPW-RC}                                                                  \\ \cline{3-8} 
\multicolumn{2}{c|}{}                          & AVG           & 100          & 240           & 500           & 640           & 900           \\ \hline
\multicolumn{2}{c|}{ $\mathcal{L}_{p}$}              & 39.8        & 9.5          & 22.4         & 45.7          & 55.6          & 69.1          \\
\multicolumn{2}{c|}{ $\mathcal{L}_{v}$}              & 39.5          & 9.4         & 22.1          & 47.0          & 57.0          & 68.9         \\
\multicolumn{2}{c|}{$\mathcal{L}_{p}+\mathcal{L}_{v}$(Ours)}                      & \textbf{39.1} & \textbf{9.2} & \textbf{21.6} & \textbf{43.6} & \textbf{52.6} & \textbf{68.5} \\ \hline
\end{tabular}

\label{ablation on loss}
\end{table}

\textbf{Effectiveness of velocity augment} We remove the input of velocity augment and utilize the original position coordinates as input to verify its validity. The experimental results in Table. \ref{velocity} demonstrates that the introduction of velocity information has advantages in modeling the joint-level motion dynamics.

\textbf{Effectiveness of intra/inter-relation learning} We compare our model with other
three settings to demonstrate the effectiveness
of intra-relation learning and inter-relation learning: {\romannumeral1}) We only perform inter-relation learning between different individuals (``w/o intra-relation'') and remove intra-relation learning; {\romannumeral2}) We only capture the individual's intra-relation (``w/o inter-relation'') and remove inter-relation learning; {\romannumeral3}) We remove the inter/intra-relation learning and adopt a 4-layer Transformer decoder (``w/o intra\&inter-relations''). It is obvious that the inter-relation learning and the intra-relation learning are both contribute to the improvement of motion prediction performance.

\textbf{Effectiveness of loss function.} 
In this section, we perform extensive
ablation studies on the 3DPW-RC datasets to investigate the
contribution of the different loss functions; see
Table \ref{ablation on loss}. We compare our strategy with other three settings: i) only $\mathcal{L}_{v}$; ii) only $ \mathcal{L}_{p}$; iii) $\mathcal{L}_{P} + \mathcal{L}_{v}$. We see that each loss function can help model get better prediction.

\textbf{Computational complexity.} 
We evaluate the trade-off
between the model’s computational cost and performance, as shown in Table. \ref{paramater}. We report the number of parameters
and an estimate of the floating operations GFLOPS (Giga Floating-point Operations Per Second) of the models when predict 14 frames (900ms) on 3DPW-RC dataset. Our proposed architecture 
achieves best performance using $L_{1}=4$ GC-blocks of intra-relation learning, $L_{2}=13$ cross-attention layers of inter-relation learning and $L_{3}=4$ aggregation layers of IAM with 5.1M parameters. Compared to JRT \cite{xu2022groupnet} and TBIFormer \cite{fang2023tbp}, we achieve better performance while using 20\% fewer parameters. Additionally, our method achieves near the best results with only 4.2M parameters.

\section{Conclusion}
In this paper, we propose a novel framework for multi-person motion prediction. It aims to avoid undesired and unnatural dependencies between persons by explicitly modeling intra-relations and inter-relations. In addition, we propose a novel plug-and-play aggregation module
called the Interaction Aggregation Module (IAM). This module
employs an aggregate-attention mechanism to seamlessly inte-
grate intra\&inter-relations. Our experiments demonstrate the strong plug-and-play capability of the IAM, and our framework's ability to accurately and naturally predict multi-person 3D motion across the 3DPW, 3DPW-RC, CMU-Mocap, MuPoTS-3D, and synthesized datasets Mix1\&Mix2 (9$\sim$15 persons). Nevertheless, our method still have some limitations. Our method mainly focuses on human-to-human interactions in the scene and lacks consideration of the impact of the environments and objects, which is inconsistent with the factors that influence a person's actions in real situations. In future work, we plan to consider the environment and objects to better comply with real-world.

\bibliographystyle{IEEEtran}
\bibliography{egbib}

\begin{thebibliography}{10}
\providecommand{\url}[1]{#1}
\csname url@rmstyle\endcsname
\providecommand{\newblock}{\relax}
\providecommand{\bibinfo}[2]{#2}
\providecommand\BIBentrySTDinterwordspacing{\spaceskip=0pt\relax}
\providecommand\BIBentryALTinterwordstretchfactor{4}
\providecommand\BIBentryALTinterwordspacing{\spaceskip=\fontdimen2\font plus
\BIBentryALTinterwordstretchfactor\fontdimen3\font minus \fontdimen4\font\relax}
\providecommand\BIBforeignlanguage[2]{{%
\expandafter\ifx\csname l@#1\endcsname\relax
\typeout{** WARNING: IEEEtran.bst: No hyphenation pattern has been}%
\typeout{** loaded for the language `#1'. Using the pattern for}%
\typeout{** the default language instead.}%
\else
\language=\csname l@#1\endcsname
\fi
#2}}

\bibitem{gopalakrishnan2019neural}
A.~Gopalakrishnan, A.~Mali, D.~Kifer, L.~Giles, and A.~G. Ororbia, ``A neural temporal model for human motion prediction,'' in \emph{Proceedings of the IEEE/CVF Conference on Computer Vision and Pattern Recognition}, 2019, pp. 12\,116--12\,125.

\bibitem{10306327}
J.~Shi, J.~Zhong, and W.~Cao, ``Multi-semantics aggregation network based on the dynamic-attention mechanism for 3d human motion prediction,'' \emph{IEEE Transactions on Multimedia}, vol.~26, pp. 5194--5206, 2024.

\bibitem{cao2022pkd}
W.~Cao, Y.~Zhang, J.~Gao, A.~Cheng, K.~Cheng, and J.~Cheng, ``Pkd: General distillation framework for object detectors via pearson correlation coefficient,'' \emph{Advances in Neural Information Processing Systems}, vol.~35, pp. 15\,394--15\,406, 2022.

\bibitem{zhu2023ipcc}
D.~Zhu, G.~Zhai, Y.~Di, F.~Manhardt, H.~Berkemeyer, T.~Tran, N.~Navab, F.~Tombari, and B.~Busam, ``Ipcc-tp: Utilizing incremental pearson correlation coefficient for joint multi-agent trajectory prediction,'' in \emph{Proceedings of the IEEE/CVF Conference on Computer Vision and Pattern Recognition}, 2023, pp. 5507--5516.

\bibitem{peng2023trajectory}
X.~Peng, S.~Mao, and Z.~Wu, ``Trajectory-aware body interaction transformer for multi-person pose forecasting,'' in \emph{Proceedings of the IEEE/CVF Conference on Computer Vision and Pattern Recognition}, 2023, pp. 17\,121--17\,130.

\bibitem{wang2023dynamic}
X.~Wang, W.~Zhang, C.~Wang, Y.~Gao, and M.~Liu, ``Dynamic dense graph convolutional network for skeleton-based human motion prediction,'' \emph{IEEE Transactions on Image Processing}, vol.~33, pp. 1--15, 2023.

\bibitem{yi2024fouriergnn}
K.~Yi, Q.~Zhang, W.~Fan, H.~He, L.~Hu, P.~Wang, N.~An, L.~Cao, and Z.~Niu, ``Fouriergnn: Rethinking multivariate time series forecasting from a pure graph perspective,'' \emph{Advances in Neural Information Processing Systems}, vol.~36, 2024.

\bibitem{zou2023snipper}
S.~Zou, Y.~Xu, C.~Li, L.~Ma, L.~Cheng, and M.~Vo, ``Snipper: A spatiotemporal transformer for simultaneous multi-person 3d pose estimation tracking and forecasting on a video snippet,'' \emph{IEEE Transactions on Circuits and Systems for Video Technology}, 2023.

\bibitem{saadat2021towards}
A.~Saadat, N.~Fathi, and S.~Saadatanejad, ``Towards human pose prediction using the encoder-decoder lstm,'' in \emph{ICCV Workshops}, 2021.

\bibitem{ioffe2015batch}
S.~Ioffe and C.~Szegedy, ``Batch normalization: Accelerating deep network training by reducing internal covariate shift,'' in \emph{International conference on machine learning}.\hskip 1em plus 0.5em minus 0.4em\relax pmlr, 2015, pp. 448--456.

\bibitem{ba2016layer}
J.~L. Ba, J.~R. Kiros, and G.~E. Hinton, ``Layer normalization,'' \emph{arXiv preprint arXiv:1607.06450}, 2016.

\bibitem{JMLR:v15:srivastava14a}
\BIBentryALTinterwordspacing
N.~Srivastava, G.~Hinton, A.~Krizhevsky, I.~Sutskever, and R.~Salakhutdinov, ``Dropout: A simple way to prevent neural networks from overfitting,'' \emph{Journal of Machine Learning Research}, vol.~15, no.~56, pp. 1929--1958, 2014. [Online]. Available: \url{http://jmlr.org/papers/v15/srivastava14a.html}
\BIBentrySTDinterwordspacing

\bibitem{9157826}
E.~Corona, A.~Pumarola, G.~Alenyà, and F.~Moreno-Noguer, ``Context-aware human motion prediction,'' in \emph{2020 IEEE/CVF Conference on Computer Vision and Pattern Recognition (CVPR)}, 2020, pp. 6990--6999.

\bibitem{NEURIPS2019_d09bf415}
V.~Kosaraju, A.~Sadeghian, R.~Mart\'{\i}n-Mart\'{\i}n, I.~Reid, H.~Rezatofighi, and S.~Savarese, ``Social-bigat: Multimodal trajectory forecasting using bicycle-gan and graph attention networks,'' in \emph{Advances in Neural Information Processing Systems}, H.~Wallach, H.~Larochelle, A.~Beygelzimer, F.~d\textquotesingle Alch\'{e}-Buc, E.~Fox, and R.~Garnett, Eds., vol.~32.\hskip 1em plus 0.5em minus 0.4em\relax Curran Associates, Inc., 2019.

\bibitem{9010834}
Y.~Huang, H.~Bi, Z.~Li, T.~Mao, and Z.~Wang, ``Stgat: Modeling spatial-temporal interactions for human trajectory prediction,'' in \emph{2019 IEEE/CVF International Conference on Computer Vision (ICCV)}, 2019, pp. 6271--6280.

\bibitem{lin2022survey}
T.~Lin, Y.~Wang, X.~Liu, and X.~Qiu, ``A survey of transformers,'' \emph{AI Open}, 2022.

\bibitem{9795092}
C.~Li, H.~Yang, and J.~Sun, ``Intention-interaction graph based hierarchical reasoning networks for human trajectory prediction,'' \emph{IEEE Transactions on Multimedia}, pp. 1--12, 2022.

\bibitem{10001762}
Z.~Li, Y.~Li, L.~Tang, T.~Zhang, and J.~Su, ``Two-person graph convolutional network for skeleton-based human interaction recognition,'' \emph{IEEE Transactions on Circuits and Systems for Video Technology}, vol.~33, no.~7, pp. 3333--3342, 2023.

\bibitem{guo2022multi}
W.~Guo, X.~Bie, X.~Alameda-Pineda, and F.~Moreno-Noguer, ``Multi-person extreme motion prediction,'' in \emph{Proceedings of the IEEE/CVF Conference on Computer Vision and Pattern Recognition}, 2022, pp. 13\,053--13\,064.

\bibitem{saadatnejad2023generic}
S.~Saadatnejad, A.~Rasekh, M.~Mofayezi, Y.~Medghalchi, S.~Rajabzadeh, T.~Mordan, and A.~Alahi, ``A generic diffusion-based approach for 3d human pose prediction in the wild,'' in \emph{2023 IEEE International Conference on Robotics and Automation (ICRA)}.\hskip 1em plus 0.5em minus 0.4em\relax IEEE, 2023, pp. 8246--8253.

\bibitem{joo2016panoptic}
H.~Joo, T.~Simon, X.~Li, H.~Liu, L.~Tan, L.~Gui, S.~Banerjee, T.~Godisart, B.~Nabbe, I.~Matthews, T.~Kanade, S.~Nobuhara, and Y.~Sheikh, ``Panoptic studio: A massively multiview system for social interaction capture,'' 2016.

\bibitem{li2022exploiting}
W.~Li, H.~Liu, R.~Ding, M.~Liu, P.~Wang, and W.~Yang, ``Exploiting temporal contexts with strided transformer for 3d human pose estimation,'' \emph{IEEE Transactions on Multimedia}, vol.~25, pp. 1282--1293, 2022.

\bibitem{jiang2020mpshape}
W.~Jiang, N.~Kolotouros, G.~Pavlakos, X.~Zhou, and K.~Daniilidis, ``Coherent reconstruction of multiple humans from a single image,'' in \emph{CVPR}, 2020.

\bibitem{unknown}
M.~Hassan, V.~Choutas, D.~Tzionas, and M.~Black, ``Resolving 3d human pose ambiguities with 3d scene constraints,'' in \emph{Proceedings of the IEEE/CVF International Conference on Computer Vision}, 2019, p. 2282– 2292.

\bibitem{liu2023temporal}
J.~Liu, X.~Wang, C.~Wang, Y.~Gao, and M.~Liu, ``Temporal decoupling graph convolutional network for skeleton-based gesture recognition,'' \emph{IEEE Transactions on Multimedia}, vol.~26, pp. 811--823, 2023.

\bibitem{6795963}
S.~Hochreiter and J.~Schmidhuber, ``Long short-term memory,'' \emph{Neural Computation}, vol.~9, no.~8, pp. 1735--1780, 1997.

\bibitem{veličković2018graph}
P.~Veličković, G.~Cucurull, A.~Casanova, A.~Romero, P.~Liò, and Y.~Bengio, ``Graph attention networks,'' 2018.

\bibitem{adeli2020socially}
V.~Adeli, E.~Adeli, I.~Reid, J.~C. Niebles, and H.~Rezatofighi, ``Socially and contextually aware human motion and pose forecasting,'' \emph{IEEE Robotics and Automation Letters}, vol.~5, no.~4, pp. 6033--6040, 2020.

\bibitem{nie2023triplet}
X.~Nie, X.~Chen, H.~Jin, Z.~Zhu, Y.~Yan, and D.~Qi, ``Triplet attention transformer for spatiotemporal predictive learning,'' \emph{arXiv preprint arXiv:2310.18698}, 2023.

\bibitem{aksan2021spatio}
E.~Aksan, M.~Kaufmann, P.~Cao, and O.~Hilliges, ``A spatio-temporal transformer for 3d human motion prediction,'' in \emph{2021 International Conference on 3D Vision (3DV)}.\hskip 1em plus 0.5em minus 0.4em\relax IEEE, 2021, pp. 565--574.

\bibitem{mao2020history}
W.~Mao, M.~Liu, and M.~Salzmann, ``History repeats itself: Human motion prediction via motion attention,'' in \emph{Computer Vision--ECCV 2020: 16th European Conference, Glasgow, UK, August 23--28, 2020, Proceedings, Part XIV 16}.\hskip 1em plus 0.5em minus 0.4em\relax Springer, 2020, pp. 474--489.

\bibitem{ghosh2017learning}
P.~Ghosh, J.~Song, E.~Aksan, and O.~Hilliges, ``Learning human motion models for long-term predictions,'' in \emph{2017 International Conference on 3D Vision (3DV)}.\hskip 1em plus 0.5em minus 0.4em\relax IEEE, 2017, pp. 458--466.

\bibitem{dang2021msr}
L.~Dang, Y.~Nie, C.~Long, Q.~Zhang, and G.~Li, ``Msr-gcn: Multi-scale residual graph convolution networks for human motion prediction,'' in \emph{Proceedings of the IEEE/CVF International Conference on Computer Vision}, 2021, pp. 11\,467--11\,476.

\bibitem{chen2019crowd}
C.~Chen, Y.~Liu, S.~Kreiss, and A.~Alahi, ``Crowd-robot interaction: Crowd-aware robot navigation with attention-based deep reinforcement learning,'' in \emph{2019 international conference on robotics and automation (ICRA)}.\hskip 1em plus 0.5em minus 0.4em\relax IEEE, 2019, pp. 6015--6022.

\bibitem{vu2020anomaly}
T.-H. Vu, S.~Ambellouis, J.~Boonaert, and A.~Taleb-Ahmed, ``Anomaly detection in surveillance videos by future appearance-motion prediction.'' in \emph{VISIGRAPP (5: VISAPP)}, 2020, pp. 484--490.

\bibitem{fang2023tbp}
S.~Fang, Z.~Wang, Y.~Zhong, J.~Ge, and S.~Chen, ``Tbp-former: Learning temporal bird's-eye-view pyramid for joint perception and prediction in vision-centric autonomous driving,'' in \emph{Proceedings of the IEEE/CVF Conference on Computer Vision and Pattern Recognition}, 2023, pp. 1368--1378.

\bibitem{xu2022groupnet}
C.~Xu, M.~Li, Z.~Ni, Y.~Zhang, and S.~Chen, ``Groupnet: Multiscale hypergraph neural networks for trajectory prediction with relational reasoning,'' in \emph{Proceedings of the IEEE/CVF Conference on Computer Vision and Pattern Recognition}, 2022, pp. 6498--6507.

\bibitem{wang2021pvred}
H.~Wang, J.~Dong, B.~Cheng, and J.~Feng, ``Pvred: A position-velocity recurrent encoder-decoder for human motion prediction,'' \emph{IEEE Transactions on Image Processing}, vol.~30, pp. 6096--6106, 2021.

\bibitem{zhong2022spatio}
C.~Zhong, L.~Hu, Z.~Zhang, Y.~Ye, and S.~Xia, ``Spatio-temporal gating-adjacency gcn for human motion prediction,'' in \emph{Proceedings of the IEEE/CVF Conference on Computer Vision and Pattern Recognition}, 2022, pp. 6447--6456.

\bibitem{sofianos2021space}
T.~Sofianos, A.~Sampieri, L.~Franco, and F.~Galasso, ``Space-time-separable graph convolutional network for pose forecasting,'' in \emph{Proceedings of the IEEE/CVF International Conference on Computer Vision}, 2021, pp. 11\,209--11\,218.

\bibitem{6682899}
C.~Ionescu, D.~Papava, V.~Olaru, and C.~Sminchisescu, ``Human3.6m: Large scale datasets and predictive methods for 3d human sensing in natural environments,'' \emph{IEEE Transactions on Pattern Analysis and Machine Intelligence}, vol.~36, no.~7, pp. 1325--1339, 2014.

\bibitem{9709907}
V.~Adeli, M.~Ehsanpour, I.~Reid, J.~C. Niebles, S.~Savarese, E.~Adeli, and H.~Rezatofighi, ``Tripod: Human trajectory and pose dynamics forecasting in the wild,'' in \emph{2021 IEEE/CVF International Conference on Computer Vision (ICCV)}, 2021, pp. 13\,370--13\,380.

\bibitem{wang2021multi}
J.~Wang, H.~Xu, M.~Narasimhan, and X.~Wang, ``Multi-person 3d motion prediction with multi-range transformers,'' \emph{Advances in Neural Information Processing Systems}, vol.~34, pp. 6036--6049, 2021.

\bibitem{vendrow2022somoformer}
E.~Vendrow, S.~Kumar, E.~Adeli, and H.~Rezatofighi, ``Somoformer: Multi-person pose forecasting with transformers,'' 2022.

\bibitem{xu2023joint}
Q.~Xu, W.~Mao, J.~Gong, C.~Xu, S.~Chen, W.~Xie, Y.~Zhang, and Y.~Wang, ``Joint-relation transformer for multi-person motion prediction,'' in \emph{Proceedings of the IEEE/CVF International Conference on Computer Vision}, 2023, pp. 9816--9826.

\bibitem{Marcard_2018_ECCV}
T.~von Marcard, R.~Henschel, M.~J. Black, B.~Rosenhahn, and G.~Pons-Moll, ``Recovering accurate 3d human pose in the wild using imus and a moving camera,'' in \emph{Proceedings of the European Conference on Computer Vision (ECCV)}, September 2018.

\bibitem{wang2021simple}
C.~Wang, Y.~Wang, Z.~Huang, and Z.~Chen, ``Simple baseline for single human motion forecasting,'' in \emph{Proceedings of the IEEE/CVF International Conference on Computer Vision}, 2021, pp. 2260--2265.

\bibitem{8460651}
J.~Bütepage, H.~Kjellström, and D.~Kragic, ``Anticipating many futures: Online human motion prediction and generation for human-robot interaction,'' in \emph{2018 IEEE International Conference on Robotics and Automation (ICRA)}, 2018, pp. 4563--4570.

\bibitem{9145701}
V.~Adeli, E.~Adeli, I.~Reid, J.~C. Niebles, and H.~Rezatofighi, ``Socially and contextually aware human motion and pose forecasting,'' \emph{IEEE Robotics and Automation Letters}, vol.~5, no.~4, pp. 6033--6040, 2020.

\bibitem{10.1007/978-3-030-58452-8_23}
Z.~Cao, H.~Gao, K.~Mangalam, Q.-Z. Cai, M.~Vo, and J.~Malik, ``Long-term human motion prediction with scene context,'' in \emph{Computer Vision -- ECCV 2020}, A.~Vedaldi, H.~Bischof, T.~Brox, and J.-M. Frahm, Eds.\hskip 1em plus 0.5em minus 0.4em\relax Cham: Springer International Publishing, 2020, pp. 387--404.

\bibitem{cmumocap2003}
``{CMU Graphics Lab Motion Capture Database},'' \url{http://mocap.cs.cmu.edu/}.

\bibitem{mehta2018single}
D.~Mehta, O.~Sotnychenko, F.~Mueller, W.~Xu, S.~Sridhar, G.~Pons-Moll, and C.~Theobalt, ``Single-shot multi-person 3d pose estimation from monocular rgb,'' in \emph{2018 International Conference on 3D Vision (3DV)}.\hskip 1em plus 0.5em minus 0.4em\relax IEEE, 2018, pp. 120--130.

\bibitem{loshchilov2017decoupled}
I.~Loshchilov and F.~Hutter, ``Decoupled weight decay regularization,'' \emph{arXiv preprint arXiv:1711.05101}, 2017.

\bibitem{wu2014leveraging}
D.~Wu and L.~Shao, ``Leveraging hierarchical parametric networks for skeletal joints based action segmentation and recognition,'' in \emph{Proceedings of the IEEE conference on computer vision and pattern recognition}, 2014, pp. 724--731.

\bibitem{10194334}
C.~Liu and Y.~Mu, ``Multi-granularity interaction for multi-person 3d motion prediction,'' \emph{IEEE Transactions on Circuits and Systems for Video Technology}, vol.~34, no.~3, pp. 1546--1558, 2024.

\bibitem{ma2022progressively}
T.~Ma, Y.~Nie, C.~Long, Q.~Zhang, and G.~Li, ``Progressively generating better initial guesses towards next stages for high-quality human motion prediction,'' in \emph{Proceedings of the IEEE/CVF Conference on Computer Vision and Pattern Recognition}, 2022, pp. 6437--6446.

\bibitem{10025861}
J.~Tang, J.~Zhang, R.~Ding, B.~Gu, and J.~Yin, ``Collaborative multi-dynamic pattern modeling for human motion prediction,'' \emph{IEEE Transactions on Circuits and Systems for Video Technology}, vol.~33, no.~8, pp. 3689--3700, 2023.

\bibitem{DBLP:journals/corr/abs-1904-03278}
\BIBentryALTinterwordspacing
N.~Mahmood, N.~Ghorbani, N.~F. Troje, G.~Pons{-}Moll, and M.~J. Black, ``{AMASS:} archive of motion capture as surface shapes,'' \emph{CoRR}, vol. abs/1904.03278, 2019. [Online]. Available: \url{http://arxiv.org/abs/1904.03278}
\BIBentrySTDinterwordspacing

\bibitem{mao2019learning}
W.~Mao, M.~Liu, M.~Salzmann, and H.~Li, ``Learning trajectory dependencies for human motion prediction,'' in \emph{Proceedings of the IEEE/CVF international conference on computer vision}, 2019, pp. 9489--9497.

\bibitem{parsaeifard2021learning}
B.~Parsaeifard, S.~Saadatnejad, Y.~Liu, T.~Mordan, and A.~Alahi, ``Learning decoupled representations for human pose forecasting,'' in \emph{Proceedings of the IEEE/CVF International Conference on Computer Vision}, 2021, pp. 2294--2303.

\bibitem{tang2023collaborative}
B.~Tang, Y.~Zhong, C.~Xu, W.-T. Wu, U.~Neumann, Y.~Zhang, S.~Chen, and Y.~Wang, ``Collaborative uncertainty benefits multi-agent multi-modal trajectory forecasting,'' \emph{IEEE Transactions on Pattern Analysis and Machine Intelligence}, 2023.

\bibitem{10064318}
H.~Yu, X.~Fan, Y.~Hou, W.~Pei, H.~Ge, X.~Yang, D.~Zhou, Q.~Zhang, and M.~Zhang, ``Toward realistic 3d human motion prediction with a spatio-temporal cross- transformer approach,'' \emph{IEEE Transactions on Circuits and Systems for Video Technology}, vol.~33, no.~10, pp. 5707--5720, 2023.

\bibitem{gao2021human}
R.~X. Gao, L.~Wang, P.~Wang, J.~Zhang, and H.~Liu, ``Human motion recognition and prediction for robot control,'' in \emph{Advanced Human-Robot Collaboration in Manufacturing}.\hskip 1em plus 0.5em minus 0.4em\relax Springer, 2021, pp. 261--282.

\bibitem{xu2022remember}
C.~Xu, W.~Mao, W.~Zhang, and S.~Chen, ``Remember intentions: Retrospective-memory-based trajectory prediction,'' in \emph{Proceedings of the IEEE/CVF Conference on Computer Vision and Pattern Recognition}, 2022, pp. 6488--6497.

\end{thebibliography}

\end{document}